\documentclass{article} 
\usepackage{iclr2026_conference,times}

\usepackage{amsmath,amsfonts,bm}

\def\eqref#1{equation~\ref{#1}}

\def\1{\bm{1}}

\DeclareMathAlphabet{\mathsfit}{\encodingdefault}{\sfdefault}{m}{sl}
\SetMathAlphabet{\mathsfit}{bold}{\encodingdefault}{\sfdefault}{bx}{n}

\usepackage{hyperref}
\usepackage{url}

\usepackage{enumitem}
\usepackage[utf8]{inputenc} 
\usepackage[T1]{fontenc}    
\usepackage{url}            
\usepackage{booktabs}       
\usepackage{amsfonts}       
\usepackage{nicefrac}       
\usepackage{microtype}      
\usepackage{xcolor}         

\usepackage{graphicx} 
\usepackage{xcolor}
\usepackage{adjustbox}
\usepackage{multirow} 
\usepackage{comment}

\usepackage[table]{xcolor}
\usepackage{booktabs}
\usepackage{multirow}
\usepackage{graphicx}
\usepackage{colortbl}
\usepackage{array}

\usepackage[table,xcdraw]{xcolor}
\usepackage{pgf}
\usepackage{colortbl}
\usepackage{booktabs}
\usepackage{multirow}

\usepackage[table,xcdraw]{xcolor}
\usepackage{pgf}

\usepackage[table,xcdraw]{xcolor}

\NewDocumentCommand{\jin}{ mO{} }{\textcolor{blue}{\textsuperscript{\textit{MJ}}\textsf{\textbf{\small[#1]}}}}

\usepackage[table,xcdraw]{xcolor}

\usepackage{amssymb}  
\usepackage{xcolor}   

\usepackage{pifont}
\usepackage{xcolor}

\definecolor{darkgreen}{RGB}{0,100,0}  

\usepackage{adjustbox}
\newcommand{\ourbenchmark}{{\sf AccidentBench}} 
\usepackage{threeparttable}

\title{AccidentBench: Benchmarking Multimodal Understanding and Reasoning in Vehicle Accidents and Beyond}

\author{%
\\
\\
  \textbf{Shangding Gu\textsuperscript{1}\thanks{Correspondence to: \textit{shangding.gu@berkeley.edu}. \\
  \hspace*{1.5em} Technical Report. Work in progress.
}},\hspace{2pt}
  \textbf{Xiaohan Wang\textsuperscript{2}},
  \textbf{Donghao Ying\textsuperscript{1}},
  \textbf{Haoyu Zhao\textsuperscript{3}},
  \textbf{Runing Yang\textsuperscript{4}}, \vspace{5pt} \\
   \textbf{Ming Jin\textsuperscript{4}},
  \textbf{Boyi Li\textsuperscript{1,5}},  
  \textbf{Marco Pavone\textsuperscript{2,5}},
  \textbf{Serena Yeung-Levy\textsuperscript{2}},
  \textbf{Jun Wang\textsuperscript{3}}, \vspace{5pt}  \\
  \textbf{Dawn Song\textsuperscript{1}},
  \textbf{Costas Spanos\textsuperscript{1}} \\ \And
  \textsuperscript{1}UC Berkeley \  
  \textsuperscript{2}Stanford \ 
  \textsuperscript{3}UCL \  
  \textsuperscript{4}Virginia Tech \  
  \textsuperscript{5}Nvidia 
}

\iclrfinalcopy 
\begin{document}

\begin{figure}[h!]
    \centering
    \includegraphics[width=0.1\linewidth]{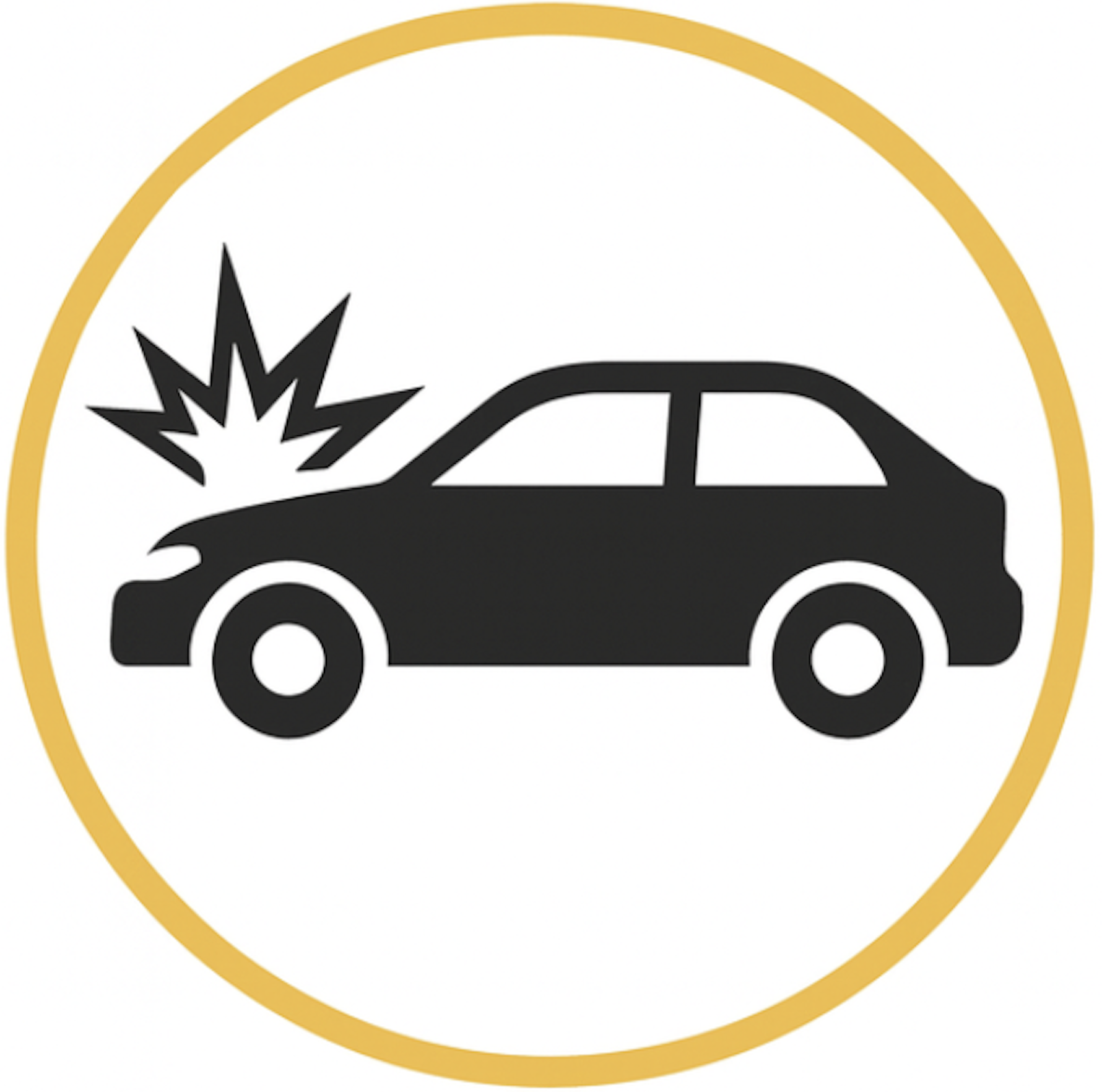}
    \label{fig:placeholder}
    \vspace{-60pt}
\end{figure}
\vspace{-22pt}
\maketitle

\begin{abstract}
Rapid advances in multimodal models demand benchmarks that rigorously evaluate understanding and reasoning in safety-critical, dynamic real-world settings. We present \ourbenchmark, a large-scale benchmark that combines \emph{vehicle accident} scenarios with \emph{Beyond} domains, safety-critical settings in air and water that emphasize spatial and temporal reasoning (e.g., navigation, orientation, multi-vehicle motion). The benchmark contains approximately 2{,}000 videos and over 19{,}000 human-annotated question--answer pairs spanning multiple video lengths (short/medium/long) and difficulty levels (easy/medium/hard). Tasks systematically probe core capabilities: temporal, spatial, and intent understanding and reasoning.  By unifying accident-centric traffic scenes with broader safety-critical scenarios in air and water, \ourbenchmark~offers a comprehensive, physically grounded testbed for evaluating models under real-world variability. Evaluations of state-of-the-art models (e.g., Gemini-2.5 Pro and GPT-5) show that even the strongest models achieve only about 18\% accuracy on the hardest tasks and longest videos, revealing substantial gaps in real-world temporal, spatial, and intent reasoning. \ourbenchmark~is designed to expose these critical gaps and drive the development of multimodal models that are safer, more robust, and better aligned with real-world safety-critical challenges. The code and dataset are available at: \url{https://github.com/SafeRL-Lab/AccidentBench} 

\end{abstract}

\section{Introduction}

As artificial intelligence (AI) continues to evolve, large multimodal models have shown impressive capabilities across vision, language, and video domains. However, significant challenges remain in deploying these models for real-world, safety-critical applications such as autonomous driving, robotics, and aerial or maritime operations. While  multimodal models demonstrate remarkable performance in constrained or simulated environments, their robustness and depth of understanding in high-stakes, dynamic scenarios are still far from sufficient.%

In particular, deployment in mission-critical domains requires rigorous evaluation of models' understanding and reasoning abilities under real-world conditions that involve uncertainty, physical interactions, and causal dependencies. While recent benchmarks have advanced evaluation in specific facets like temporal understanding (e.g., MVBench \citep{li2024mvbench}, REXTIME \citep{chen2024rextime}) or domain-specific knowledge (e.g., MMMU \citep{yue2024mmmu}, DriveLM \citep{sima2024drivelm}), there remains a paucity of unified platforms that assess understanding and reasoning across diverse vehicle accident and other open-space domains.
To address this, we designed \ourbenchmark~to rigorously evaluate multimodal models’ understanding and reasoning in safety-critical tasks, with a primary focus on traffic accident scenarios and other high-stakes real-world settings.

\begin{figure}
    \centering
    \includegraphics[width=1.0\linewidth]{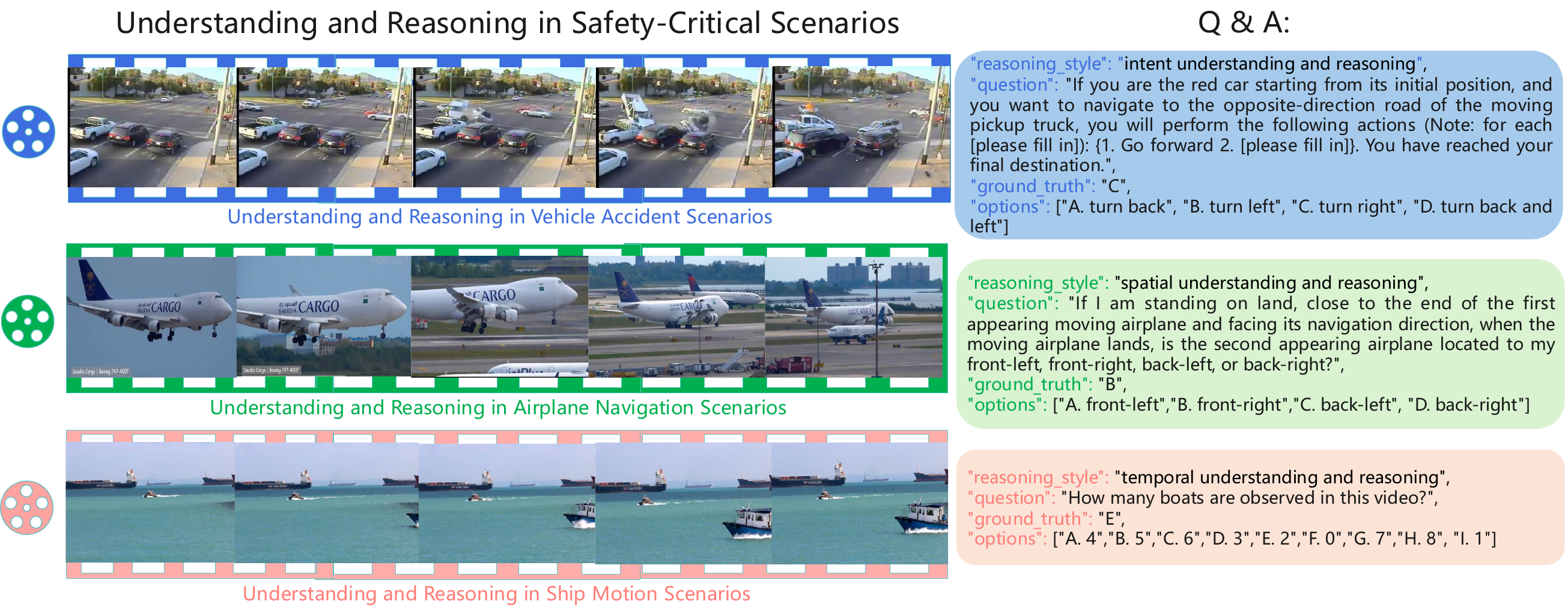}
    \caption{Examples of multimodal understanding and reasoning in vehicle accident and other safety-critical scenarios.}
    \label{fig:overview-open-space-QA}
    \vspace{-15pt}
\end{figure}

Specifically, \ourbenchmark~targets understanding and reasoning across diverse vehicle accident scenarios (83.0\%), while also encompassing airspace (10.2\%) and waterway (6.8\%) domains, in which safety, perception, and decision-making are deeply interdependent. Unlike benchmarks that emphasize isolated skills or single domains, \ourbenchmark~systematically challenges models across several critical understanding and reasoning capabilities: {temporal understanding and reasoning} (tracking event sequences and causality over extended periods); {spatial understanding and reasoning} (understanding dynamic spatial relationships and multi-agent trajectories); and {intent and goal reasoning} (inferring agent intentions and planning goals), which further includes {complex strategic and counterfactual reasoning} (evaluating higher-order strategies, action implications, and ``what-if'' scenarios). Representative examples from \ourbenchmark~are illustrated in Figure~\ref{fig:overview-open-space-QA}. By probing these abilities across diverse, safety-critical scenarios, \ourbenchmark~offers a rigorous framework for assessing progress toward multimodal AI systems capable of reliable real-world operation.

{Our key contributions are summarized as follows:}

\begin{itemize}[leftmargin=*]      
    \item \textbf{Vehicle Accident Focus:} We introduce \ourbenchmark, which emphasizes diverse vehicle accident scenarios while also extending to airspace and waterway domains. Evaluating vehicle accidents is especially critical for the safe deployment of LLMs in real-world applications and is a key step toward their widespread use in autonomous driving.  
    \item \textbf{Real-World Limitations and Safety Gaps:} We highlight weaknesses in current AI systems’ understanding and reasoning across open-space domains (e.g., autonomous driving, aviation, and marine) and provide a challenging testbed to advance safer and more reliable multimodal models.    
    \item \textbf{Unified Evaluation Suite:} \ourbenchmark~is a large-scale, video-based benchmark that integrates land traffic, airspace, and waterway scenarios, systematically evaluating temporal understanding, spatial understanding, and intent/goal reasoning within dynamic, safety-critical environments.  
\end{itemize}

\section{Related Work}

\subsection{General Multimodal Understanding Benchmarks}

Recent years have witnessed growing interest in video understanding benchmarks. Foundational video question-answering (QA) efforts include MSR-VTT~\citep{xu2016msr} and Next-QA~\citep{xiao2021next}. More recently, MVBench~\citep{li2024mvbench}, with its 20 diverse temporal tasks derived from static images, and MLVU~\citep{zhou2024mlvu} have expanded video QA capabilities across multiple domains. The challenge of long-form video understanding has seen contributions from benchmarks such as EgoSchema~\citep{mangalam2023egoschema}, Video-LLaVA~\citep{fu2024video}, MovieChat~\citep{song2024moviechat}, and LongVideoBench~\citep{wu2024longvideobench}. Parallelly, video captioning benchmarks such as AuroraCap~\citep{chai2024auroracap}, HiCM2~\citep{kim2025hicm2}, and LongCaptioning~\citep{wei2025longcaptioning} focus on generating detailed textual descriptions.

A significant trend is the push for more rigorous temporal and causal reasoning.  REXTIME~\citep{chen2024rextime}, for instance, probes the linking of causally related events across separate video segments. For multi-domain understanding, MMWorld~\citep{he2025mmworld} evaluates models across diverse disciplines, requiring explanations and counterfactuals. Furthermore, LVBench~\citep{wang2024lvbench} integrates video inputs for QA. Beyond video, reasoning from static images is explored by MME~\citep{jiang2025mme} (including CoT extensions), MMMU~\citep{yue2024mmmu} (evaluating expert-level multi-discipline reasoning), and benchmarks for mathematical reasoning like Dynamath~\citep{zou2024dynamath} and MultiModal-MATH~\citep{zhou2024multimodal}. For academic content, Video-MMLU~\citep{song2025video} offers a large-scale lecture video benchmark.


While these diverse benchmarks advance important aspects of multimodal understanding, such as general video comprehension, temporal analysis, long-form narrative understanding, captioning, and static image reasoning, they typically lack a unified framework for evaluation across land, air, and maritime open-space environments. Moreover, they may not capture the specific combination of complex reasoning skills, including strategic and intent-based inference, that \ourbenchmark~is designed to assess in these contexts.

\subsection{Safety-Critical  Multimodal  Understanding Benchmarks}

Evaluating models in safety-critical domains, where understanding and reasoning under uncertainty is vital, is an emerging focus. Initial efforts addressed static image safety~\citep{liu2024mm}, model robustness against adversarial attacks (e.g., FigStep~\citep{gong2023figstep}, JailBreakV~\citep{luo2024jailbreakv})~\citep{shayegani2023jailbreak, qi2024visual}, or indoor robotics~\citep{yang2024thinking}. 

Autonomous driving has been a major driver of safety-critical research. Foundational datasets such as nuScenes\footnote{\url{https://www.nuscenes.org/}} and Waymo Open Dataset\footnote{\url{https://waymo.com/open/}}, along with language-integrated efforts such as DriveLM and DriveVLM~\citep{sima2024drivelm,tian2025drivevlm}, are closely related to \ourbenchmark's goals due to their real-world video and safety considerations. However, a key motivation for \ourbenchmark~is that these traditionally emphasized perception and planning, with less focus on deep safety-critical reasoning for tasks such as accident cause analysis or complex decision-making.  Other specialized benchmarks tackle related issues such as video anomaly detection (e.g., VANE-Bench~\citep{gani2025vane}).


While advancements continue in specialized video reasoning and domain-specific safety evaluations, existing benchmarks still largely focus on single operational domains. Critically, they often lack sufficient coverage of high-risk scenarios such as traffic collisions, ship navigation, and airplane takeoff/landing events across combined land, air, and water settings. A unified platform to consistently evaluate robust, generalizable reasoning (e.g., temporal-causal, spatial, intent, and strategic analysis) across these diverse, safety-critical real-world scenarios also remains absent. To address this specific void, \ourbenchmark~distinctively incorporates these challenging high-risk scenarios from all three domains. The reliability of its complex reasoning evaluation is ensured as all annotations were generated by highly educated annotators. \ourbenchmark~thus provides a much-needed testbed for fostering robust, adaptable AI capable of safety-critical scenario understanding.

\section{Benchmark Design and Analysis}

\subsection{Scenario Settings}

In this benchmark, we include diverse real-world scenario datasets \footnote{These datasets are used solely for academic research. They are employed only to evaluate model performance in this work.}, with a primary focus on traffic accident understanding and reasoning. {Vehicle accident} scenarios account for 83\% of the dataset. In addition, we incorporate high-stakes, safety-critical settings such as {airplane navigation} scenarios, which account for 10.8\% and focus on takeoff and landing, and {ship motion} scenarios, which account for 6.2\% and emphasize navigation understanding and reasoning.

\paragraph{Vehicle Accident Scenarios}

In the scenarios, we include a comprehensive range of traffic accident scenarios, encompassing diverse collision events under varying weather conditions such as snow, rain, and sunshine, as detailed in Table~\ref{table:land-space-traffic-elements}. Specific examples of these scenarios are illustrated in Figure~\ref{fig:land-space-traffic-accident-understanding-reasoning}, and more detailed examples are provided in Appendix \ref{appendix:details-examples}. To enhance contextual diversity, we incorporate multiple camera perspectives, including ego-centric and third-person views, particularly for accident scenes. The dataset features incidents involving a wide variety of vehicle types, including buses, motorcycles, sedans, and several categories of trucks, across different road environments such as highways, freeways, and rural roads. The associated questions are designed to evaluate models across multiple reasoning dimensions, including temporal-causal understanding, spatial reasoning, and intent and goal planning. The original video datasets are sourced from \citep{BaoMM2020, shah2018cadp}, which primarily collected videos from YouTube and other public internet platforms.

\begin{figure}
    \centering
    \includegraphics[width=1.0\linewidth]{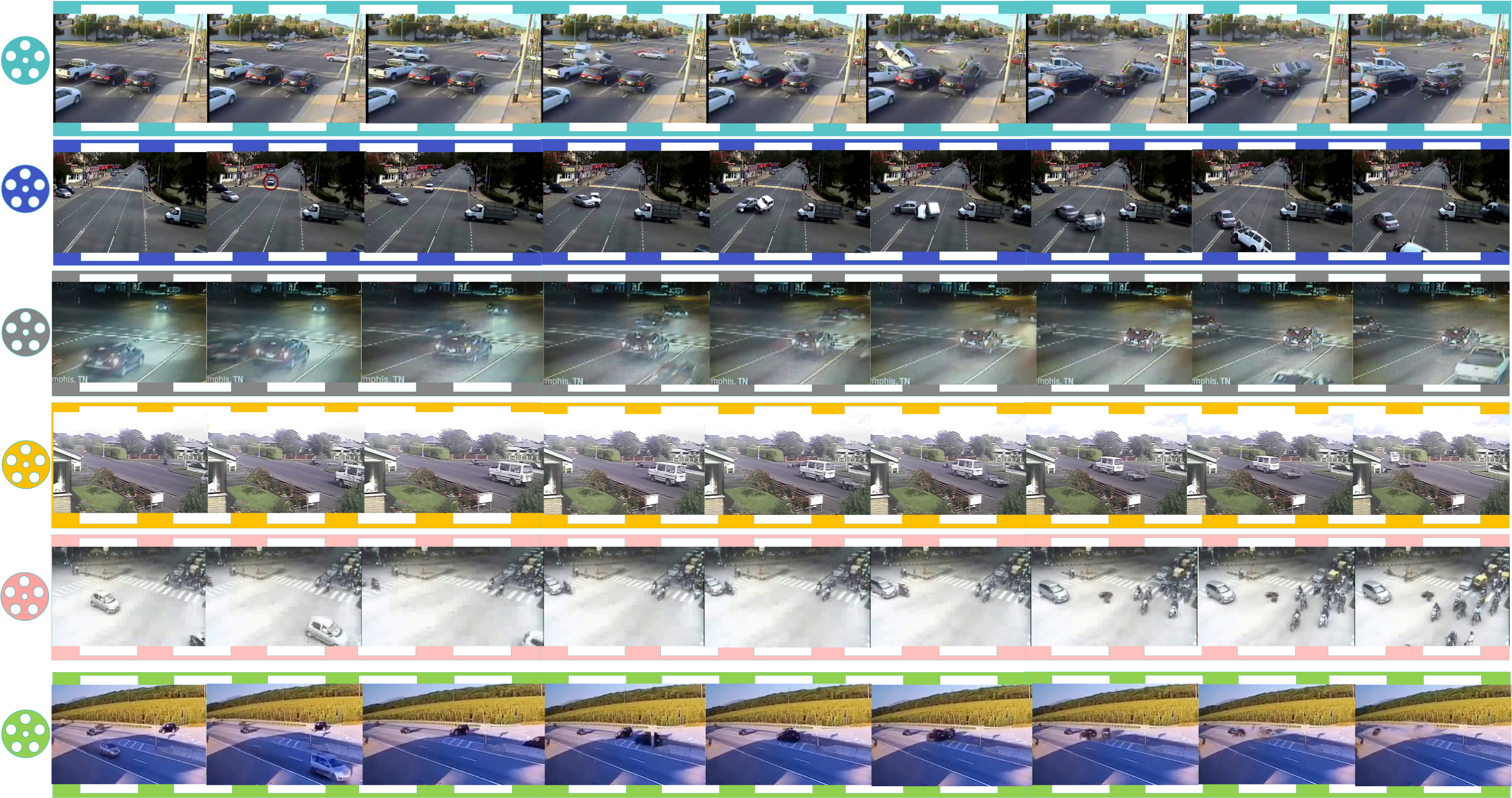}
    \caption{Land-space traffic accident scenarios for open-space video understanding and reasoning include \textcolor{cyan}{intersection collisions}, \textcolor{blue}{ urban road accidents}, \textcolor{gray}{nighttime incidents}, \textcolor{orange}{rural road accidents}, \textcolor{pink}{snow-covered road collisions}, and \textcolor{green}{freeway accidents}.}
    \label{fig:land-space-traffic-accident-understanding-reasoning}
\end{figure}

\begin{table}[t]
\centering
\caption{Overview of traffic accident scenarios in our benchmark, covering diverse road environments, weather conditions, and involved traffic participants.}
\scriptsize
\begin{adjustbox}{width=1\textwidth,center}
\begin{tabular}{ll}
\hline
\textbf{Index} & \textbf{Categories} \\
\hline
\textbf{Road Environments:} & Intersection, Highway, Freeway, Rural Road, Tunnel, Urban Road, Bridge,  Parking Lot \\
\textbf{Weather Conditions:} & Snow, Rain, Sunshine, Cloudy, Foggy, Windy \\
\textbf{Involved Participants:} & Sedan, SUV, Bus, Truck, Motorcycle, Bicycle, Van, Pickup, Trailer, Pedestrian \\
\hline
\end{tabular}
\end{adjustbox}
\label{table:land-space-traffic-elements}
\vspace{-15pt}
\end{table}

\paragraph{Other Safety-Critical Scenarios}
\textit{(1) Ship Motion Scenarios:} These scenarios include both \textbf{river} and \textbf{ocean} settings, covering diverse boats and ships under varying navigation conditions. These environments are critical yet underexplored in multimodal research. We assess temporal, spatial, and intent/goal understanding and reasoning through video-based tasks of different lengths and difficulty levels, using both interval-based and accuracy-based formats. The water-space videos are sourced from publicly available datasets, including \citep{guo2023asynchronous, prasad2017video}.
\textit{(2) Airplane Navigation Scenarios:} These scenarios primarily involve {takeoff} and {landing} events, emphasizing airplane navigation and perceptual understanding and reasoning. Despite their real-world importance, airplanes also remain underexplored in multimodal research. Our benchmark captures variations in navigation patterns, aircraft sizes, and motion dynamics across different airplane types. These scenarios include videos of varying lengths and evaluate models on spatial, temporal, and intent/goal understanding and reasoning across multiple difficulty levels using both interval-based and accuracy-based multiple-choice formats. The airspace videos are sourced from publicly available footage\footnote{\url{https://www.youtube.com/watch?v=i6CrbqeksJ8}},\footnote{\url{https://www.youtube.com/watch?v=k5yvzTw08K8}},\footnote{\url{https://www.youtube.com/watch?v=Bt9tpiAmTs8}}.

\begin{figure}[tb!]
    \centering
\includegraphics[width=1.0\linewidth]{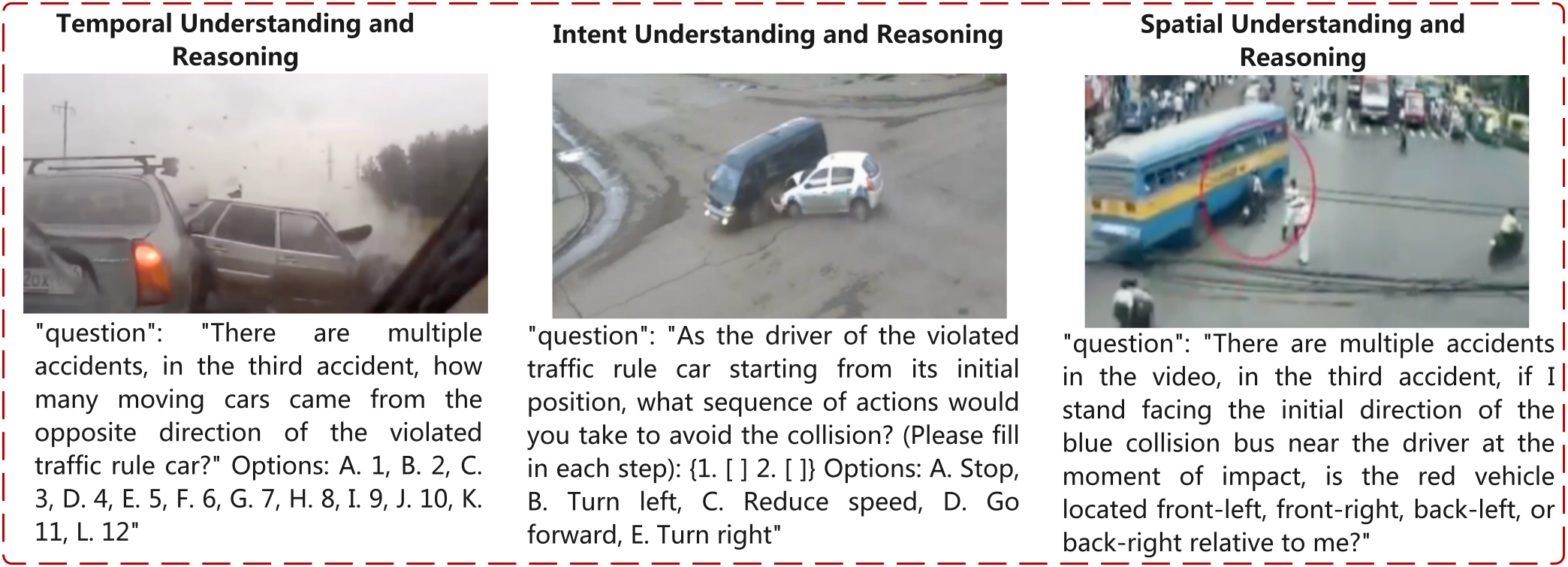}
\caption{Examples of question settings in \ourbenchmark~across three key understanding and reasoning types: \textit{Temporal Understanding and Reasoning}, which involves understanding event sequences and motion over time; \textit{Spatial Understanding and Reasoning}, which focuses on relative positioning and orientation in space; and \textit{Intent Understanding and Reasoning}, which evaluates understanding of goal-directed behaviors and decision-making in dynamic environments.}
    \label{fig:three-reasoning-styles-examples}
\end{figure}

\subsection{Task Settings}

Within each scenario, we design tasks that evaluate models across three key dimensions of understanding and reasoning: \textit{temporal}, \textit{spatial}, and \textit{intent and goal}. Representative examples for each type are shown in Figure~\ref{fig:three-reasoning-styles-examples}.

For each understanding and reasoning dimension, we construct tasks at three difficulty levels using two formats: \textit{interval-based choices} (easy and medium) and \textit{accuracy-based choices} (hard). Easy tasks ($\approx$6,300 QA pairs) provide approximately three coarse-grained interval options; medium tasks ($\approx$6,300 QA pairs) include six intermediate-level intervals; and hard tasks ($\approx$6,300 QA pairs) present fine-grained discrete options that require an exact match with the correct answer. The number of tasks is evenly distributed across difficulty levels, with each tier comprising one-third of the total. In all cases, the model must select a single best answer, allowing the benchmark to systematically assess performance under increasing levels of precision and ambiguity.

\subsection{Dataset Analysis}

This benchmark includes approximately 2,000 videos and related massive human-annotated question-answer pairs, covering a wide range of understanding and reasoning tasks. The dataset features a variety of video lengths, categories, and frame counts, and spans real-world scenarios. An overview of the dataset’s characteristics is provided in Appendix~\ref{appendix:details-examples}, which illustrates the distributions of video duration, domain coverage, and task styles, along with details of the annotation procedure and difficulty levels.

\subsection{Comparison with Existing Benchmarks}
Table \ref{tab:benchmark} provides a comparative analysis of \ourbenchmark~alongside existing evaluation benchmarks for multimodal models. Most benchmarks primarily focus on assessing the multimodal reasoning capabilities of multimodal models \citep{he2024mmworld,song2023moviechat,zhou2024mlvu}; however, a significant limitation is the prevalent oversight of safety considerations. While a few recent benchmarks have begun to evaluate safety aspects of multimodal models \citep{zhou2024multimodal,liu2024mm}, they typically do not incorporate video-based question answering and are mostly limited to single-frame inputs. However, single-frame capture often introduces uncertainties in reasoning and is insufficient for reliably assessing multimodal models’ ability to handle safety-critical issues.
 In contrast, our \ourbenchmark~introduces a large-scale curated collection of video question-answer pairs that specifically focus on traffic accident understanding and reasoning in real-world safety-related scenarios. Comprising 2,000 carefully selected videos and 6,300 hard-level question--answer pairs, extended to medium and easy levels by varying the number of answer choices, \ourbenchmark~includes a total of over 19,000 question--answer pairs. This scale is competitive with existing benchmarks and highlights the comprehensiveness of our evaluation set.

\begin{table*}[tb!]
\centering
\renewcommand{\arraystretch}{0.92}
\caption{\textbf{Benchmark comparison} for multimodal understanding and reasoning tasks.}
\resizebox{\textwidth}{!}{%
\begin{tabular}{l| cccc | ccc | cc }
\toprule
\multirow{2}{*}{\textbf{Dataset}} & \multirow{2}{*}{\textbf{Safety}}& \multirow{2}{*}{\textbf{Traffic}} & \multirow{2}{*}{\textbf{Annotation}} & \multirow{2}{*}{\textbf{Real-World}} & \multirow{2}{*}{\textbf{Main Scenarios}}& \multirow{2}{*}{\textbf{\#~Video}} & \multirow{2}{*}{\textbf{Ave. Duration (s)}} & \multicolumn{2}{c}{\textbf{Question-answering}} \\
& & & & & & & & \textbf{Number} & \textbf{Type} \\
\midrule
MovieChat-1K~\citep{song2023moviechat} & \textcolor{red}{\textbf{\texttimes}} & \textcolor{red}{\textbf{\texttimes}} & Human & \textcolor{darkgreen}{\checkmark} & General & 1,000 & 564 & 13,000 & Open-ended \\
MMWorld~\citep{he2024mmworld} & \textcolor{red}{\textbf{\texttimes}} & \textcolor{red}{\textbf{\texttimes}} & Human & \textcolor{darkgreen}{\checkmark} & General & 1,910 & 107 & 6,627 & Multiple-choice \\
MLVU~\citep{zhou2024mlvu} & \textcolor{red}{\textbf{\texttimes}} & \textcolor{red}{\textbf{\texttimes}} & Human & \textcolor{darkgreen}{\checkmark} & General & 1,730 & 930 & 3,102 & Multiple-choice \\
MVBench~\citep{abellan2023probing} & \textcolor{red}{\textbf{\texttimes}} & \textcolor{red}{\textbf{\texttimes}} & Human \& LLM & \textcolor{darkgreen}{\checkmark} & General & 4,000 & 16 & 4,000 & Multiple-choice \\
LongVideoBench~\citep{wu2024longvideobench} & \textcolor{red}{\textbf{\texttimes}} & \textcolor{red}{\textbf{\texttimes}} & Human & \textcolor{darkgreen}{\checkmark} & General &  3,763 & 473 & 6,678 & Multiple-choice \\
TempCompass~\citep{liu2024tempcompass} & \textcolor{red}{\textbf{\texttimes}} & \textcolor{red}{\textbf{\texttimes}} & Human \& LLM & \textcolor{darkgreen}{\checkmark} & General & 410 & $<30$ & 7,540 & Multiple-choice\\
VSI-Bench \citep{yang2024thinking} & \textcolor{red}{\textbf{\texttimes}} & \textcolor{red}{\textbf{\texttimes}} & Human & \textcolor{darkgreen}{\checkmark} & Embodied & 288 & 50-100 & 5,000 & Multiple-choice \\
Video-MMMU~\citep{hu2025video} & \textcolor{red}{\textbf{\texttimes}} & \textcolor{red}{\textbf{\texttimes}} & Human \& LLM & \textcolor{red}{\textbf{\texttimes}} & Professional & 300 & 506 & 900 & Multiple-choice \\
Video-MMLU \citep{song2025video} & \textcolor{red}{\textbf{\texttimes}} & \textcolor{red}{\textbf{\texttimes}} & Human \& LLM & \textcolor{red}{\textbf{\texttimes}} & Professional & 1,065 & 109 & 15,746 & Open-ended \\
DriveBench \citep{xie2025vlms} & \textcolor{darkgreen}{\checkmark} & \textcolor{darkgreen}{\checkmark} & Human \& LLM & \textcolor{darkgreen}{\checkmark} & General Driving & \textcolor{red}{\textbf{\texttimes}} &\textcolor{red}{\textbf{\texttimes}} & 19,200 & Multiple-choice \\
DriveLM \citep{sima2023drivelm} & \textcolor{darkgreen}{\checkmark} & \textcolor{darkgreen}{\checkmark} & Human & \textcolor{darkgreen}{\checkmark} & General Driving & \textcolor{red}{\textbf{\texttimes}} &\textcolor{red}{\textbf{\texttimes}} & 15,480 & Open-ended \\
nuScenes-QA \citep{qian2024nuscenes} & \textcolor{red}{\textbf{\texttimes}} & \textcolor{darkgreen}{\checkmark} & Human & \textcolor{darkgreen}{\checkmark} & General Driving & \textcolor{red}{\textbf{\texttimes}} &\textcolor{red}{\textbf{\texttimes}} & 83,337 & Open-ended \\
MSSBench \citep{zhou2024multimodal} & \textcolor{darkgreen}{\checkmark} & \textcolor{red}{\textbf{\texttimes}} & Human \& LLM & \textcolor{darkgreen}{\checkmark} & General & \textcolor{red}{\textbf{\texttimes}} & \textcolor{red}{\textbf{\texttimes}} & 1960 & Open-ended\\
MMSBench \citep{liu2024mm} & \textcolor{darkgreen}{\checkmark} & \textcolor{red}{\textbf{\texttimes}} & LLM & \textcolor{darkgreen}{\checkmark} & General & \textcolor{red}{\textbf{\texttimes}} &\textcolor{red}{\textbf{\texttimes}} & 5040 & Open-ended \\
\midrule
\ourbenchmark~(ours) & \textcolor{darkgreen}{\checkmark} & \textcolor{darkgreen}{\checkmark} & Human & \textcolor{darkgreen}{\checkmark} & Accident & 2000 & 56 & 19,000 & Multiple-choice \\
\bottomrule
\end{tabular}%
}
\label{tab:benchmark}
\end{table*}
\section{Experiments}

In our experiments, we build upon the \texttt{lmms-eval} framework~\citep{zhang2024lmmsevalrealitycheckevaluation} as the foundation for our benchmark and extend it to support the specific requirements of \ourbenchmark. We conduct comprehensive evaluations to assess the performance of state-of-the-art (SOTA) multimodal models across diverse safety-critical real-world scenarios.

\begin{table*}[tb!]
\centering 
\caption{Understanding and reasoning evaluation for \ourbenchmark~in Vehicle Accidents.}
\begin{adjustbox}{width=0.99\textwidth,center}
\begin{tabular}{ll|cccc|cccc|cccc}
\toprule
\multirow{2}{*}{Models} & \multirow{2}{*}{Size}
& \multicolumn{4}{c|}{\textbf{Hard}}
& \multicolumn{4}{c|}{\textbf{Medium}}
& \multicolumn{4}{c}{\textbf{Easy}} \\
\cmidrule(lr){3-6}\cmidrule(lr){7-10}\cmidrule(lr){11-14}
& & Avg. & Temp. & Spatial & Intent
  & Avg. & Temp. & Spatial & Intent
  & Avg. & Temp. & Spatial & Intent \\
\midrule
GPT 5 \citep{chatgpt5-2025openai}                & -   & \textbf{37.33}  & 35.85 & 42.80 & 33.35  & \textbf{48.34}  & 46.22 & 55.07 & 43.74 & {54.86}  & 52.35 & 55.50 & 56.72 \\
GPT 4o~\citep{hurst2024gpt}                 & -   & 25.82 & 29.61 & 31.38 & 13.21 & {43.05} & 47.63 & 48.59 & 31.83 & 44.17 & 54.45 & 36.01 & 43.67 \\
Gemini 2.5 pro~\citep{gemini2dot5-2025}     & -   & {31.06} & 38.75 & 37.54 & 23.46 & 40.57 & 39.13 & {47.22} & 30.33 & \textbf{57.90} & 58.24 & 56.23 & 55.52 \\
Gemini 2.5 flash think~\citep{gemini2025flash} & - & 29.90 & 34.52 & 36.57 & 23.00 & 39.50 & 45.18 & 45.61 & 29.76 & 48.93 & 58.35 & 51.21 & 37.78 \\
Gemini 2.5 flash no-think~\citep{gemini2025flash} & - & 23.80 & 24.43 & 33.04 & 17.55 & 36.67 & 41.21 & 43.86 & 25.44 & 46.89 & 50.92 & 50.24 & 35.19 \\
Gemini 1.5 pro~\citep{google2024gemini15}   & -   & 17.79 & 20.90 & 20.72 & 15.81 & 35.98 & 39.05 & 41.11 & 28.75 & 47.00 & 56.01 & 45.68 & 40.25 \\
Claude 3.5~\citep{anthropic-claude-2025}    & -   & 30.82 & 35.04 & 31.65 & 22.91 & 37.93 & 36.39 & 46.36 & {32.63} & 51.08 & 53.32 & 47.01 & 48.93 \\
InternVL2.5~\citep{chen2024expanding}       & 26B & 23.92 & 31.00 & 29.75 & 11.50 & 35.42 & 41.75 & 43.00 & 22.75 & 56.33 & 61.00 & 56.25 & 46.50 \\
InternVL2.5~\citep{chen2024expanding}       & 8B  & 21.25 & 24.50 & 31.25 & 10.50 & 34.83 & 42.25 & 48.25 & 14.50 & 52.34 & 55.50 & 57.00 & 42.50 \\
InternVL2.5~\citep{chen2024expanding}       & 4B  & 17.50 & 19.50 & 25.50 & 12.00 & 35.33 & 34.00 & 41.25 & 26.50 & 48.00 & 46.00 & 51.50 & 43.50 \\
LLaVA Next~\citep{li2024llavanext-ablations} & 32B & 19.34 & 13.50 & 24.50 & 11.00 & 21.83 & 15.50 & 31.25 & 14.00 & 37.09 & 27.25 & 41.75 & 35.00 \\
LLaVA Video~\citep{zhang2024video}          & 7B  & 19.67 & 15.00 & 31.25 & 12.00 & 25.42 & 22.00 & 32.50 & 22.50 & 30.58 & 31.00 & 32.25 & 34.00 \\
LLaVA OneVision~\citep{li2024llava}         & 7B  & 13.83 & 8.50 & 21.75 & 12.00 & 16.67 & 20.50 & 19.00 & 17.00 & 30.83 & 29.75 & 32.25 & 29.00 \\
Qwen2.5 VL~\citep{Qwen2.5_VL}               & 32B & 23.33 & 18.00 & 29.50 & 18.00 & 27.99 & 25.75 & 38.50 & 23.50 & 45.67 & 53.00 & 45.50 & 41.25 \\
Qwen2.5 VL~\citep{Qwen2.5_VL}               & 7B  & 23.42 & 18.25 & 30.50 & 20.75 & 32.17 & 30.50 & 37.00 & 24.00 & 43.58 & 44.00 & 38.50 & 42.75 \\
\bottomrule
\end{tabular}
\end{adjustbox}
\label{tab:land-space-weighted}
\end{table*}

\subsection{Evaluation in Vehicle Accident Scenarios}

We evaluate model performance across all vehicle accident scenarios in \ourbenchmark, with results summarized in Table~\ref{tab:land-space-weighted}. The evaluation is organized by task difficulty (Easy, Medium, Hard) and reasoning type (Temporal, Spatial, Intent). Among the models, \textbf{GPT-5} achieves the strongest overall performance, leading in the Hard setting with an average score of 37.33 and maintaining high results in Medium (48.34). \textbf{Gemini~2.5 Pro} also performs consistently well, ranking best in the Easy setting (57.90) and remaining competitive in Medium (40.57) and Hard (31.06). \textbf{GPT-4o} shows strong results in Medium (43.05) and Easy (44.17) tasks, particularly in temporal and spatial reasoning, but its performance drops sharply on Hard tasks (25.82). \textit{Across all models, performance declines substantially as task difficulty increases, with intent reasoning under the Hard setting posing the most difficult challenge.} Overall, proprietary models (e.g., GPT-5, Gemini, GPT-4o) outperform open-source counterparts, but none achieves robust performance across all difficulty levels and reasoning types.

\subsection{Vehicle Accident Evaluation Analysis} 

To investigate how video length and task format affect model performance in vehicle accident scenarios, we report results from accuracy-based (hard) experiments and interval-based (easy and medium) experiments across short, medium, and long video lengths.

\paragraph{Accuracy-Based Settings} 
As shown in Table~\ref{tab:land-space-short-medium-long-accuracy}, we present a comprehensive evaluation of model performance in the \textbf{Vehicle Accident} scenarios of \ourbenchmark, categorized by task type, video length. In the hard (accuracy-based) setting, performance drops significantly across all models as video length increases. For example, in hard tasks involving long videos, \textit{even the best-performing models fall below 40\% average accuracy and only achieve around 18\% accuracy on the hardest tasks and longest video scenarios.} These results highlight the limitations of current multimodal models in handling complex, long-horizon real-world understanding and reasoning—particularly for extended temporal sequences, fine-grained spatial relations, and intent understanding and reasoning.

\begin{table*}[tb!]
\centering
\renewcommand{\arraystretch}{1.1}
\vspace{-15pt}
\caption{Evaluation of \ourbenchmark~on \textbf{vehicle accident} scenarios using \textbf{short}, \textbf{medium}, and \textbf{long} videos, categorized by reasoning types and based on a subset of the dataset. The choices are \textbf{accuracy-based}, corresponding to the hard setting.}

\Large{
	\begin{adjustbox}{width=1.0\textwidth,center}
\begin{tabular}{llcc|cccc|cccc|cccc}
\toprule
\multirow{2}{*}{\textbf{Difficulty}}  & \multirow{2}{*}{\textbf{Models}}  & \multirow{2}{*}{ \textbf{Size}} &\multirow{2}{*}{ \textbf{Over. Avg.}}  
& \multicolumn{4}{c|}{\textbf{Short Video Scenarios}} & \multicolumn{4}{c|}{\textbf{Medium Video Scenarios}} & \multicolumn{4}{c}{\textbf{Long Video Scenarios}}  \\
 & & &
& \textbf{Avg.} & \textbf{Temporal} & \textbf{Spatial} & \textbf{Intent} & \textbf{Avg.} & \textbf{Temporal} & \textbf{Spatial} & \textbf{Intent} & \textbf{Avg.} & \textbf{Temporal} & \textbf{Spatial} & \textbf{Intent} \\
  \midrule
  & GPT 5 \citep{chatgpt5-2025openai}  & -   & \textbf{37.33}  & \textbf{45.87} & 48.52 & 55.10 & 34.00 & \textbf{48.12} & 49.02 & 39.29 & 56.06 & 18.00 & 10.00 & 34.00 & 10.00 \\
  & GPT 4o \citep{hurst2024gpt}  & -   & {24.41}  & {26.78} & 34.65 & 34.69 & 11.00 & {35.70} & 43.14 & 32.14 & 31.82 & 11.00 & 6.00 & 26.00 & 1.00 \\
   & Gemini 2.5 pro  \citep{gemini2dot5-2025}                  & -   &  {29.76} & {34.84} & 36.63 & 44.90 & 23.00  & {35.76} & 45.10 & 30.36 & 31.82 & \textbf{18.67} & 10.00 & 28.00 & 18.0\\
  & Gemini 2.5 flash think  \citep{gemini2025flash}                  & -   & 28.67 & 32.13 & 35.64 & 37.75 & 23.00 & 35.20 & 37.25 & 41.07 & 27.27 & \textbf{18.67} & 6.00 & 36.00 & 14.00 \\
  & Gemini 2.5 flash no-think  \citep{gemini2025flash}                    & -   & 24.34 & 24.74 & 30.69 & 26.53 & 17.00 & 30.94 & 52.94 & 23.21 & 16.67 & 17.33 & 14.00 & 24.00 & 14.00 \\
  & Gemini 1.5 pro \citep{google2024gemini15}  & - & 18.76  & 19.72  & 23.76 & 20.41 & 15.00 & 24.55 & 33.33 & 16.07 & 24.24 & 12.00  & 2.00   & 26.00  & 8.00 \\
  & Claude 3.5  \citep{anthropic-claude-2025} & -   & 28.71 & 33.76 & 35.64 & 31.63 & 34.00 & 28.87 & 37.26 & 35.71 & 13.63 & 16.00 & 12.00 & 26.00 & 10.0\\
   Hard   & InternVL2.5   \citep{chen2024expanding}          & 26B   & {23.78} & 21.33 & 26.00 & 31.00 &  7.00    & 32.00  & 46.00  & 32.00  & 18.00   & 18.00 & 16.00  & 24.00  & 14.00  \\
   & InternVL2.5   \citep{chen2024expanding}          & 8B    & 22.67      & 20.00 & 18.00 & 33.00 &  9.00    & 30.00  & 46.00  & 30.00  & 14.00   & 18.00 & 16.00  & 28.00  & 10.00  \\
   & InternVL2.5 \citep{chen2024expanding}            & 4B    & 19.56      & 18.67 & 18.00 & 28.00 &  8.00    & 28.00  & 34.00  & 24.00  & 26.00   & 12.00 &  8.00  & 22.00  &  6.00  \\
   & LLaVA Next~\citep{li2024llavanext-ablations} & 32B & 16.22     & 20.67 & 16.00 & 32.00 & 14.00     & 11.33     & 12.00    & 12.00   & 10.00     & 16.67    & 10.00    & 30.00    & 10.00    \\
   & LLaVA Video~\citep{zhang2024video}        & 7B    & 19.78      & 19.33 & 12.00 & 35.00 & 11.00    & 24.67  & 26.00  & 30.00  & 18.00   & 15.33 & 10.00  & 28.00  &  8.00  \\
   & LLaVA OneVision~\citep{li2024llava}       & 7B    & 13.67      & 14.33 &  5.00 & 27.00 & 11.00    & 14.67  & 18.00  &  8.00  & 18.00   & 12.00 &  6.00  & 22.00  &  8.00  \\
    & Qwen2.5 VL~\citep{Qwen2.5_VL}            & 32B   & 22.66      & 19.33 & 11.00 & 34.00 & 13.00    & 35.33  & 46.00  & 24.00  & 36.00   & 13.33 &  4.00  & 26.00  & 10.00  \\
    & Qwen2.5 VL~\citep{Qwen2.5_VL}            & 7B    & 22.89      & 26.00 & 17.00 & 30.00 & 31.00    & 30.00  & 40.00  & 32.00  & 18.00   & 12.67 &  2.00  & 30.00  &  6.00  \\

\bottomrule
\end{tabular}
\end{adjustbox}
}
\label{tab:land-space-short-medium-long-accuracy}
\end{table*}

\begin{table*}[tb!]
\centering
\renewcommand{\arraystretch}{1.1}
\vspace{-15pt}
\caption{Evaluation of \ourbenchmark~on \textbf{vehicle accident} scenarios using \textbf{short}, \textbf{medium}, and \textbf{long} videos, categorized by reasoning types and based on a subset of the dataset. The tasks use \textbf{interval-based} choices, corresponding to the easy and medium settings depending on the number of options.}

\Large{
	\begin{adjustbox}{width=1.0\textwidth,center}
\begin{tabular}{llcc|cccc|cccc|cccc}
\toprule
\multirow{2}{*}{\textbf{Difficulty}}  & \multirow{2}{*}{\textbf{Models}}  & \multirow{2}{*}{ \textbf{Size}} &\multirow{2}{*}{ \textbf{Over. Avg.}}  
& \multicolumn{4}{c|}{\textbf{Short Video Scenarios}} & \multicolumn{4}{c|}{\textbf{Medium Video Scenarios}} & \multicolumn{4}{c}{\textbf{Long Video Scenarios}}  \\
 & & &
& \textbf{Avg.} & \textbf{Temporal} & \textbf{Spatial} & \textbf{Intent} & \textbf{Avg.} & \textbf{Temporal} & \textbf{Spatial} & \textbf{Intent} & \textbf{Avg.} & \textbf{Temporal} & \textbf{Spatial} & \textbf{Intent} \\
  \midrule
  & GPT 5 \citep{chatgpt5-2025openai}  & -   & \textbf{48.34}  & {62.55} & 64.65 & 67.00 & 56.00 & {46.48} & 50.00 & 42.22 & 47.22 & {36.00} & 24.00 & 56.00 & 28.00 \\
    & GPT 4o \citep{hurst2024gpt}  & -   & {36.99}  & {45.49} & 48.48 & 55.00 & 33.00 & 33.89 & 41.67 & 26.67 & 33.33 & 31.33 & 24.00 & 44.00 & 26.00 \\
    & Gemini 2.5 pro   \citep{gemini2dot5-2025}                  & -   & {36.46} & {42.79} & 38.38 & 59.00 & 31.00 & {33.93} & 39.58 & 28.89 & 33.33 & 32.67 & 28.00 & 44.00 & 26.0\\
    & Gemini 2.5 flash think  \citep{gemini2025flash}                    & -   & {37.52} & 47.82 & 46.47 & 56.00 & 41.00 & 36.99 & 43.75 & 42.22 & 25.00 & 28.00 & 12.00 & 44.00 & 28.00\\
    & Gemini 2.5 flash no-think  \citep{gemini2025flash}                    & -   & 36.70 & 47.50 & 48.49 & 58.00 & 36.00 & 33.93 & 39.58 & 28.89 & 33.33 & 28.67 & 24.00 & 42.00 & 20.00 \\
    & Gemini 1.5 pro \citep{google2024gemini15}  & -   & 33.89  & 39.47 & 42.42 & 42.00 & 34.00 & 33.52 & 33.33 & 42.22 & 25 & 28.67 & 12.00   & 52.00  & 22.00 \\
    & Claude 3.5  \citep{anthropic-claude-2025} & -   & 35.35 & 41.78 & 35.35 & 50.00 & 40.00 & 35.60 & 39.58 & 42.22 & 25.00 & 28.67 & 16.00 & 44.00 & 26.0\\
    Medium  & InternVL2.5   \citep{chen2024expanding}                  & 26B   & {35.11}  & 36.00 & 39.00 & 50.00 & 19.00 & {36.67} & 50.00 & 36.00 & 24.00 & {32.67} & 30.00 & 40.00 & 28.00 \\
    & InternVL2.5   \citep{chen2024expanding}                  & 8B    & 34.66  & 37.33 & 43.00 & 57.00 & 12.00 & 35.33 & 42.00 & 46.00 & 18.00 & 31.33 & 26.00 & 44.00 & 24.00 \\
    & InternVL2.5  \citep{chen2024expanding}                   & 4B    & 33.89  & 39.67 & 38.00 & 53.00 & 28.00 & 32.67 & 44.00 & 28.00 & 26.00 & 29.33 & 16.00 & 46.00 & 26.00 \\
    & LLaVA Next~\citep{li2024llavanext-ablations} & 32B & 20.00     & 27.33 & 16.00 & 49.00 & 17.00 & 10.67    & 14.00   & 10.00   & 8.00   & 22.00    & 16.00   & 36.00   & 14.00   \\
    & LLaVA Video~\citep{zhang2024video}   & 7B    & 25.67  & 25.00 & 20.00 & 34.00 & 26.00 & 28.67 & 36.00 & 28.00 & 22.00 & 23.33 & 14.00 & 40.00 & 16.00 \\
    & LLaVA OneVision~\citep{li2024llava} & 7B    & 16.67  & 16.00 & 26.00 & 30.00 & 16.00 & 14.67 & 18.00 &  8.00 & 18.00 & 19.33 &  12.00 & 30.00 &  16.00 \\
    & Qwen2.5 VL~\citep{Qwen2.5_VL}   & 32B   & 28.55  & 28.33 & 21.00 & 44.00 & 20.00 & 33.33 & 40.00 & 30.00 & 30.00 & 24.00 &  8.00 & 40.00 & 24.00 \\
    & Qwen2.5 VL~\citep{Qwen2.5_VL}   & 7B    & 29.89  & 39.00 & 37.00 & 42.00 & 38.00 & 30.67 & 32.00 & 40.00 & 20.00 & 20.00 & 16.00 & 26.00 & 18.00 \\  
\midrule
 & GPT 5 \citep{chatgpt5-2025openai}  & -   & \textbf{54.86}  & {71.20} & 76.00 & 69.61 & 68.00 & {48.71} & 47.06 & 44.90 & 54.17 & {44.67} & 34.00 & 52.00 & 48.00 \\
    & GPT 4o \citep{hurst2024gpt}  & -   & 42.17  & 52.35 & 59.00 & 47.06 & 51.00 & {47.16} & 54.9 & 44.9 & 41.67 &  27.00 & 44.00 & 5.00 & 32.00 \\
     & Gemini 2.5 pro   \citep{gemini2dot5-2025}                  & -  & {54.56} & {62.96} & 70.00 & 55.88 & 63.00 & {54.73} & 52.94 & 59.18 & 52.08 & {46.00} & 40.00 & 54.00 & 44.00 \\
     & Gemini 2.5 flash think  \citep{gemini2025flash}                    & -   & 50.00 & 67.56 & 69.00 & 65.69 & 68.00 & 44.45 & 52.94 & 40.82 & 39.58 & 38.00 & 32.00 & 38.00 & 44.00\\
      & Gemini 2.5 flash no-think  \citep{gemini2025flash}                    & -   & 51.40 & 58.97 & 70.00 & 54.90 & 52.00 & 46.56 & 52.94 & 36.74 & 50.00 & 48.67 & 38.00 & 56.00 & 52.00 \\
    & Gemini 1.5 pro \citep{google2024gemini15}  & - & 46.00  & 51.33  & 60.00 & 50.00 & 44.00 & 36.92 & 49.02 & 36.73 & 25.00 & 50.00  & 58.00   & 44.00  & 48.00 \\
    & Claude 3.5  \citep{anthropic-claude-2025} & -   & 48.59 & 60.33 & 61.00 & 50.00 & 70.00 & 36.35 & 35.29 & 51.02 & 22.73 & 49.33 & 64.00 & 44.00 & 40.0\\
    Easy  & InternVL2.5   \citep{chen2024expanding}                  & 26B   & {52.55}  & 61.00 & 62.00 & 59.00 & 62.00 & 45.33 & 58.00 & 44.00 & 34.00 & {51.33} & 62.00 & 62.00 & 30.00 \\
    & InternVL2.5 \citep{chen2024expanding}                    & 8B    & 50.11  & 55.67 & 55.00 & 60.00 & 52.00 & 44.67 & 58.00 & 42.00 & 34.00 & 50.00 & 54.00 & 64.00 & 32.00 \\
     & InternVL2.5 \citep{chen2024expanding}                    & 4B    & 44.89  & 53.33 & 46.00 & 60.00 & 54.00 & 37.33 & 48.00 & 38.00 & 26.00 & 44.00 & 44.00 & 48.00 & 40.00 \\
   & LLaVA Next~\citep{li2024llavanext-ablations} & 32B   & 31.25     & 38.00 & 35.00 & 45.00 & 34.00 & 21.33    & 12.00   & 14.00   & 38.00   & 34.67    & 20.00   & 50.00   & 34.00   \\
    & LLaVA Video~\citep{zhang2024video}   & 7B    & 31.44  & 33.00 & 30.00 & 31.00 & 38.00 & 33.33 & 38.00 & 36.00 & 26.00 & 28.00 & 16.00 & 32.00 & 36.00 \\
    & LLaVA OneVision~\citep{li2024llava} & 7B    & 29.78  & 32.00 & 31.00 & 33.00 & 32.00 & 24.00 & 26.00 & 30.00 & 16.00 & 33.33 & 28.00 & 36.00 & 36.00 \\
    & Qwen2.5 VL~\citep{Qwen2.5_VL}   & 32B   & 43.22  & 51.00 & 58.00 & 50.00 & 45.00 & 41.33 & 46.00 & 38.00 & 40.00 & 37.33 & 32.00 & 44.00 & 36.00 \\
    & Qwen2.5 VL~\citep{Qwen2.5_VL}   & 7B    & 40.67  & 51.33 & 55.00 & 42.00 & 57.00 & 36.00 & 32.00 & 42.00 & 34.00 & 34.67 & 34.00 & 28.00 & 42.00 \\

\bottomrule
\end{tabular}
\end{adjustbox}
}
\label{tab:land-space-short-medium-long-interval}
\end{table*}

\paragraph{Interval-Based Settings} 
As shown in Table~\ref{tab:land-space-short-medium-long-interval}, in the easy and medium (interval-based) settings, \textbf{GPT-5} achieves the strongest overall performance, reaching 54.86\% accuracy, followed closely by \textbf{Gemini~2.5 Pro} at 54.56\%. Other proprietary models, such as Gemini~2.5 flash and GPT-4o, also perform competitively, with GPT-4o attaining 52.39\% overall accuracy. Among open-source systems, {InternVL2.5 (26B)} is the best performer, with an overall accuracy of 52.55\%. While models like Gemini~2.5 flash (with think mode) and GPT-4o achieve relatively strong results on medium-difficulty tasks (37.53\% and 36.99\%, respectively), \textit{{performance consistently declines as video length increases, highlighting the persistent challenges in achieving robust understanding and reasoning across diverse real-world scenarios.}}

\subsection{Other Open-Space Evaluation}

Beyond vehicle accident evaluation, we also assess models in other high-stakes, safety-critical scenarios (17\%), including \textbf{ship motion} (6.8\%) and \textbf{airplane navigation} (10.2\%).  \textbf{Evaluation in Ship Motion Scenarios:} Table~\ref{tab:water-space-weighted} shows results for multimodal models in the Water Space domain of \ourbenchmark, categorized by task difficulty (Easy, Medium, Hard) and reasoning type (Temporal, Spatial, Intent). \textbf{GPT-5} achieves the highest overall performance, leading in Hard (38.36), Medium (51.80), and Easy (63.00) tasks. \textbf{Gemini~2.5 Pro} remains competitive, with strong results on Hard tasks (28.11) and particularly strong spatial and intent reasoning. \textbf{Gemini~2.5 flash with think} also performs well, achieving the best results among proprietary models in Medium (46.72) and Easy (62.01) settings before GPT-5. Among open-source models, {InternVL2.5 (26B)} and {Qwen2.5} show competitive performance, especially in temporal reasoning, but still lag behind proprietary models. As with other domains, all models suffer a marked drop in performance on Hard tasks, most notably in intent reasoning. These findings emphasize the continued difficulty of multimodal reasoning in dynamic and ambiguous environments such as rivers and oceans, highlighting the need for more advanced AI systems. Due to space constraints, further analysis of ship motion across different video lengths and task modes, as well as the \textbf{Evaluation of Airplane Navigation Scenarios}, is provided in Appendix~\ref{appendix:air-space-eva}.

\begin{table*}[tb!]
\vspace{-15pt}
\centering 
\caption{Understanding and reasoning evaluation for \ourbenchmark~in ship motion scenarios.}
\begin{adjustbox}{width=0.99\textwidth,center}
\begin{tabular}{ll|cccc|cccc|cccc}
\toprule
\multirow{2}{*}{Models} & \multirow{2}{*}{Size} 
& \multicolumn{4}{c|}{\textbf{Hard}} 
& \multicolumn{4}{c|}{\textbf{Medium}} 
& \multicolumn{4}{c}{\textbf{Easy}} \\
\cmidrule(lr){3-6}\cmidrule(lr){7-10}\cmidrule(lr){11-14}
& & Avg. & Temp. & Spatial & Intent 
  & Avg. & Temp. & Spatial & Intent 
  & Avg. & Temp. & Spatial & Intent \\
\midrule
GPT 5 \citep{chatgpt5-2025openai}                & -   & \textbf{38.36}  & 38.08 & 31.39 & 45.62  & \textbf{51.80}  & 54.93 & 47.08 & 53.39 & \textbf{63.00}  & 69.77 & 49.00 & 70.23 \\
GPT4o~\citep{hurst2024gpt}                 & -   & 19.97 & 22.62 & 21.29 & 14.73  & 37.30 & 39.48 & 50.53 & 20.84  & 47.16 & 63.06 & 38.68 & 41.69 \\
Gemini 2.5 pro~\citep{gemini2dot5-2025}    & -   & {28.11} & 33.38 & 22.06 & 26.80 & 40.92 & 44.30 & 56.28 & 29.45 & 60.92 & 68.47 & 52.68 & 58.04 \\
Gemini 2.5 flash think~\citep{gemini2025flash}    & -   & 27.17 & 31.39 & 29.15 & 23.42 & {46.72} & 52.02 & 56.24 & 35.00 & {62.01} & 65.20 & {74.48} & 52.74 \\
Gemini 2.5 flash no-think~\citep{gemini2025flash} & -   & 24.76 & 25.61 & {38.20} & 18.02 & 42.27 & 42.37 & 50.28 & 32.30 & 59.15 & 57.84 & 68.01 & 51.06 \\
Gemini 1.5 pro~\citep{google2024gemini15}  & -   & 25.48 & 31.25 & 23.57 & 21.97 & 41.86 & {48.64} & 50.01 & {41.17} & 49.84 & 47.47 & 50.30 & 50.02 \\
Claude 3.5~\citep{anthropic-claude-2025}   & -   & 24.14 & 23.67 & 20.77 & 26.06 & 39.26 & 40.07 & 53.80 & 26.67 & 50.27 & 58.37 & 52.46 & 39.70 \\
InternVL2.5~\citep{chen2024expanding}      & 26B & 22.35 & 17.68 & 25.19 & 22.01 & 41.01 & 25.68 & 60.34 & 34.78 & 52.42 & 55.55 & 51.60 & 43.28 \\
InternVL2.5~\citep{chen2024expanding}      & 8B  & 21.98 & 13.74 & 27.65 & 21.21 & 41.01 & 33.81 & 60.90 & 25.26 & 51.51 & 57.54 & 51.19 & 46.09 \\
InternVL2.5~\citep{chen2024expanding}      & 4B  & 20.92 & 17.01 & 24.68 & 21.60 & 44.13 & 27.18 & 62.23 & 44.04 & 53.28 & 52.10 & 55.76 & 44.42 \\
LLaVA Next~\citep{li2024llavanext-ablations} & 32B & 13.85 & 7.96 & 27.13 & 7.98 & 20.18 & 10.84 & 33.10 & 16.68 & 35.00 & 34.48 & 39.46 & 33.38 \\
LLaVA Video~\citep{zhang2024video}         & 7B  & 13.45 & 9.70 & 21.59 & 7.10 & 22.14 & 19.81 & 29.13 & 18.95 & 30.31 & 23.56 & 37.22 & 30.00 \\
LLaVA OneVision~\citep{li2024llava}        & 7B  & 15.00 & 9.42 & 27.25 & 8.42 & 22.59 & 16.27 & 32.09 & 18.29 & 32.95 & 29.67 & 37.08 & 31.44 \\
Qwen2.5 VL~\citep{Qwen2.5_VL}              & 32B & 12.99 & 7.97 & 23.63 & 7.37 & 33.25 & 19.69 & 50.00 & 29.72 & 52.04 & 45.12 & 56.49 & 43.05 \\
Qwen2.5 VL~\citep{Qwen2.5_VL}              & 7B  & 13.76 & 7.02 & 26.33 & 8.00 & 26.10 & 18.94 & 28.36 & 24.67 & 30.17 & 34.70 & 20.74 & 34.95 \\
\bottomrule
\end{tabular}
\end{adjustbox}
\label{tab:water-space-weighted}
\end{table*}

These findings demonstrate \ourbenchmark's ability to {reveal the limitations of existing multimodal models, particularly in safety-critical and physically grounded domains.} \textit{By highlighting domain-specific understanding and reasoning gaps, especially in underexplored high-stakes environments such as ship motion, and airplane navigation,} \ourbenchmark~serves as a useful tool for guiding the development of more robust, spatially, temporally aware, and {intent-aware} multimodal systems.

\subsection{Model Error Analysis}

\begin{figure}[h!]
    \centering
    \vspace{-10pt}
    \includegraphics[width=0.95\linewidth]{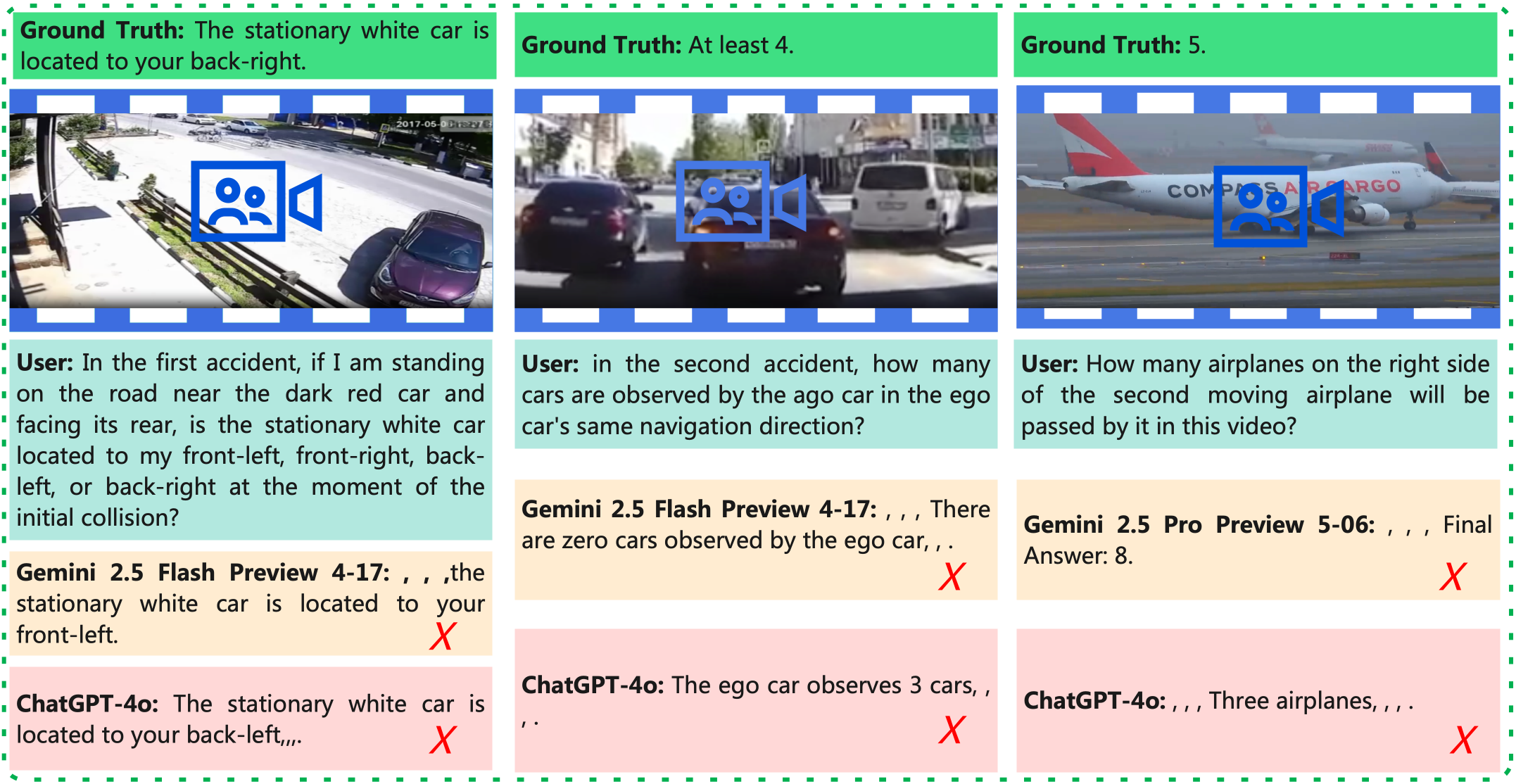}
    \caption{Qualitative error analysis of SOTA multimodal models (Gemini 2.5 and GPT-4o) on the \ourbenchmark~benchmark. Each example illustrates a failure case in a different reasoning category: spatial reasoning (left), temporal reasoning (middle), and intent reasoning (right). Despite their capabilities, both models struggle with spatial localization, counting dynamic objects, and understanding goal-directed motion in real-world safety-critical scenarios.}

    \label{fig:model-error-analysis}
\end{figure}

To demonstrate the effectiveness of our benchmark and evaluate the performance of SOTA models, we conduct a qualitative analysis of model predictions on the \ourbenchmark~benchmark. As shown in Figure~\ref{fig:model-error-analysis}, the analysis highlights persistent challenges in spatial, temporal, and intent understanding and reasoning across real-world environments. Despite the strong overall performance of leading multimodal models such as Gemini 2.5 and GPT-4o, the results reveal consistent failure cases in real-world scenarios. For example, both models struggle with accurately identifying spatial relationships (e.g., relative positions of vehicles), counting dynamic objects over time (e.g., cars in motion), and understanding goal-directed interactions (e.g., airplane passing events). \textit{These failure cases highlight the limitations of current models in handling safety-critical, perception-intensive tasks.} By providing richly annotated, video-based tasks that demand multi-step reasoning grounded in physics, causality, and spatial understanding, \ourbenchmark~serves as a rigorous diagnostic benchmark. Our findings highlight the necessity of such benchmarks for advancing the robustness, safety, and real-world applicability of large multimodal systems.

\subsection{Ablation Experiments}

In our experiments, due to the high cost of evaluating all data points, we adopt a uniform sampling strategy to select a representative subset of tasks. Specifically, for each understanding and reasoning type, we sample 50 tasks when the total number of available tasks is fewer than 500, and 100 tasks when the number exceeds 500. The \ourbenchmark~spans three real-world scenarios, {vehicle accident}, {airplane navigation}, and {ship motion}, each with three video lengths (short, medium, long), three difficulty levels (easy, medium, hard), and three understanding and reasoning types: temporal, spatial, and intent-based understanding and reasoning. Following this sampling strategy, we evaluate a total of 3,798 tasks, evenly distributed across the three types: 1,266 \textit{spatial understanding and reasoning}, 1,266 \textit{temporal-causal 
 understanding and reasoning}, and 1,266 \textit{intent understanding and reasoning} tasks.  To assess the reliability of this sampling approach, we conduct an ablation study comparing model performance on sampled tasks versus the full set of data points in the \textbf{vehicle accident (short, easy)} settings. We use InternVL 2.5, one of the leading open-source multimodal models, which ranks highly on several leaderboards such as \footnote{\url{https://enxinsong.com/Video-MMLU-web/}} and \footnote{\url{https://huggingface.co/spaces/opencompass/open_vlm_leaderboard}}. As shown in Table~\ref{table:abaltion-sample-full-datapoints}, performance on the sampled subset is comparable to, and in some cases slightly better than, performance on the full dataset. These results validate the effectiveness of our sampling strategy in preserving benchmark consistency while reducing evaluation cost.

\begin{table}[ht]
\centering
\renewcommand{\arraystretch}{1.2}
\caption{Performance comparison on \textbf{vehicle accident short videos} (easy setting): full vs. sampled data points.}
\begin{adjustbox}{width=0.7\textwidth,center}
\begin{tabular}{l|cccc|cccc}
\toprule
\multirow{2}{*}{\textbf{Model}} & \multicolumn{4}{c|}{\textbf{Full Data Points}} & \multicolumn{4}{c}{\textbf{Sample Data Points}} \\
 & \textbf{Avg.} & \textbf{Temporal} & \textbf{Spatial} & \textbf{Intent} & \textbf{Avg.} & \textbf{Temporal} & \textbf{Spatial} & \textbf{Intent} \\
\midrule
{InternVL2\_5-26B} & 55.62 & 57.61 & 50.37 & 58.88 & 61.00 & 62.00 & 59.00 & 62.00 \\
{InternVL2\_5-8B}  & 49.26 & 51.89 & 48.57 & 47.31 & 55.67 & 55.00 & 60.00 & 52.00 \\
{InternVL2\_5-4B}  & 50.65 & 50.17 & 50.70 & 51.10 & 55.33 & 52.00 & 55.00 & 59.00 \\
\bottomrule
\end{tabular}
\end{adjustbox}
\label{table:abaltion-sample-full-datapoints}
\end{table}

\section{Conclusion}

In this work, we introduce \ourbenchmark, a large-scale benchmark for evaluating multimodal understanding and reasoning in real-world safety-scitical environments. \ourbenchmark~ provides richly annotated, video-based tasks designed to assess model performance across three fundamental understanding and reasoning dimensions: temporal, spatial, and intent and goal reasoning. The benchmark encompasses a broad range of scenarios, video lengths, and difficulty levels, enabling comprehensive evaluation in safety-critical, perception-intensive settings. Through extensive qualitative and quantitative analyses, we demonstrate that even SOTA multimodal models, both proprietary systems such as Gemini 2.5 Pro and GPT-5, and leading open-source models like Qwen and InternVL, exhibit significant limitations when understanding and reasoning over complex, dynamic physical environments. We hope that \ourbenchmark~will serve as a valuable resource for the research community and help advance the development of safer, more generalizable, and practically deployable multimodal AI systems.

\bibliography{sources/reference}
\bibliographystyle{iclr2026_conference}


\clearpage

\textbf{\Large Appendix}
\appendix

\section{Limitation and Impact}

\paragraph{Limitation}
 Our benchmark provides a valuable tool for evaluating model performance in safety-critical environments. However, due to the large scale of the dataset, evaluating all data points is computationally expensive. As a result, we were unable to perform large-scale testing with many high-cost proprietary models such as ChatGPT and Gemini. In future work, we plan to explore more efficient evaluation strategies and extend our analysis to a broader set of models, including closed-source systems.
\paragraph{Impact}
 This benchmark offers a new direction for advancing multimodal model development in open-space, safety-critical, and physically grounded real-world environments. By emphasizing temporal, spatial, and intent-based reasoning in diverse video scenarios, this benchmark can be useful to guide the design of more robust and reliable multimodal systems. While this research seeks to advance the capabilities of AI in complex settings, we do not identify any specific societal risks or consequences requiring special attention at this time.

\section{Air Space Evaluation:}
\label{appendix:air-space-eva}

Table \ref{tab:air-space-weighted} reports the evaluation results for multimodal models in the airplane navigation of \ourbenchmark. The results are broken down by task difficulty (Easy, Medium, Hard) and reasoning types (Temporal, Spatial, Intent). Gemini 2.5 Pro stands out with the strongest overall performance, achieving the highest average scores across all difficulty levels, including 31.86 (Hard), 41.21 (Medium), and 55.74 (Easy). It particularly excels in intent reasoning, reaching up to 61.17 in the Easy setting. GPT-5 and GPT-4o also perform competitively, for example, GPT-4o achieve good results on Easy tasks (40.72) and intent reasoning (39.67), though it lags behind Gemini on harder examples. Open-source models such as InternVL2.5 and Qwen2.5 show moderate success in temporal reasoning but consistently underperform in intent reasoning. Overall, the trend mirrors that of the Land domain: performance declines significantly as difficulty increases, with the largest drop occurring in temporal and intent reasoning tasks. These results emphasize the challenges multimodal models face in reliably operating in dynamic, real-world Air Space scenarios.
using a Short:Medium:Long video length weighting of 0.4167:0.4167:0.1667.

\begin{table*}[ht]
\centering\caption{Understanding and reasoning evaluation for \ourbenchmark~in Airplane Navigation domain.}
\begin{adjustbox}{width=0.99\textwidth,center}
\begin{tabular}{ll|cccc|cccc|cccc}
\toprule
\multirow{2}{*}{Models} & \multirow{2}{*}{Size}
& \multicolumn{4}{c|}{\textbf{Hard}}
& \multicolumn{4}{c|}{\textbf{Medium}}
& \multicolumn{4}{c}{\textbf{Easy}} \\
\cmidrule(lr){3-6}\cmidrule(lr){7-10}\cmidrule(lr){11-14}
& & Avg. & Temp. & Spatial & Intent
  & Avg. & Temp. & Spatial & Intent
  & Avg. & Temp. & Spatial & Intent \\
\midrule
GPT 5 \citep{chatgpt5-2025openai}                & -   & {28.11}  & 26.67 & 28.00 & 29.67  & {44.00}  & 43.00 & 44.33 & 44.67 & {52.00}  & 51.00 & 41.00 & 64.00 \\
GPT 4o~\citep{hurst2024gpt} & -
    & 18.02 & 12.21 & 29.77 & 15.46
    & 30.53 & 31.33 & 40.83 & 31.83
    & 40.72 & 37.83 & 37.00 & 39.67 \\
Gemini 2.5 pro~\citep{gemini2dot5-2025} & -
    & \textbf{31.86} & 34.26 & 21.56 & 34.25
    & 41.21 & 44.08 & 38.25 & 53.50
    & \textbf{55.74} & 59.72 & 47.17 & 61.17 \\
Gemini 2.5 flash think~\citep{gemini2025flash} & -
    & 25.78 & 26.00 & 18.00 & 34.00
    & 39.78 & 39.33 & 32.00 & 48.00
    & 50.67 & 49.33 & 40.00 & 62.00 \\
Gemini 2.5 flash no-think~\citep{gemini2025flash} & -
    & 25.44 & 25.33 & 22.00 & 28.00
    & \textbf{49.67} & 43.33 & 30.00 & 52.00
    & 50.78 & 49.33 & 36.00 & 60.00 \\
Gemini 1.5 pro~\citep{google2024gemini15} & -
    & 22.88 & 19.15 & 24.75 & 22.25
    & 36.21 & 32.83 & 49.50 & 32.00
    & 43.89 & 40.56 & 41.89 & 49.67 \\
Claude 3.5~\citep{anthropic-claude-2025} & -
    & 24.31 & 16.55 & 32.30 & 23.00
    & 36.44 & 32.60 & 47.79 & 33.33
    & 41.03 & 37.56 & 41.61 & 45.33 \\
InternVL2.5~\citep{chen2024expanding} & 26B
    & 18.60 & 17.75 & 26.50 & 12.00
    & 20.14 & 26.31 & 23.31 & 46.83
    & 32.11 & 36.31 & 34.25 & 46.42 \\
InternVL2.5~\citep{chen2024expanding} & 8B
    & 18.71 & 14.80 & 29.75 & 10.00
    & 23.73 & 30.42 & 32.92 & 46.00
    & 37.86 & 40.33 & 36.50 & 40.00 \\
InternVL2.5~\citep{chen2024expanding} & 4B
    & 15.14 & 14.25 & 16.75 & 13.13
    & 24.41 & 27.00 & 28.25 & 46.75
    & 38.31 & 39.64 & 38.39 & 41.25 \\
LLaVA Next~\citep{li2024llavanext-ablations} & 32B
    & 18.23 & 8.98 & 35.08 & 10.15
    & 20.71 & 17.47 & 37.33 & 21.67
    & 28.60 & 32.69 & 34.36 & 34.67 \\
LLaVA Video~\citep{zhang2024video} & 7B
    & 15.56 & 8.48 & 25.80 & 9.00
    & 20.35 & 20.25 & 25.83 & 21.33
    & 29.62 & 30.94 & 30.97 & 30.00 \\
LLaVA OneVision~\citep{li2024llava} & 7B
    & 15.76 & 11.00 & 26.75 & 9.50
    & 19.81 & 19.84 & 23.83 & 20.83
    & 29.62 & 30.94 & 30.97 & 30.00 \\
Qwen2.5 VL~\citep{Qwen2.5_VL} & 32B
    & 16.35 & 3.43 & 31.08 & 13.75
    & 35.85 & 29.00 & 27.17 & 43.67
    & 51.73 & 52.33 & 40.61 & 54.44 \\
Qwen2.5 VL~\citep{Qwen2.5_VL} & 7B
    & 16.38 & 1.16 & 30.00 & 16.00
    & 28.70 & 22.61 & 30.33 & 25.83
    & 38.92 & 35.78 & 36.39 & 30.00 \\
\bottomrule
\end{tabular}
\end{adjustbox}
\label{tab:air-space-weighted}
\end{table*}

\section{Air and Water Space Analysis:} Table~\ref{tab:air-space-short-medium-long} presents model performance in the \textbf{Airplane Navigation} of \ourbenchmark, evaluated across short, medium, and long video scenarios, and categorized by temporal, spatial, and intent reasoning tasks. In the easy setting, \textbf{Gemini 2.5 Pro} achieves the highest overall accuracy (52.56\%), outperforming all other models, including GPT-4o and GPT-5. In the medium setting, Gemini 2.5 flash without think mode leads with 49.67\%, followed closely by GPT-5 (44.00\%) and Gemini Pro(43.11\%). For hard tasks, which are the most challenging, \textbf{Gemini 2.5 Pro} remains the top performer with 31.39\%. These results highlight the ability of the Gemini family of models to maintain performance in complex, dynamic airspace environments, but exhibit notable drops as the reasoning complexity increases, revealing current limitations in handling temporal, spatial, and intent-based challenges in aerial domains. Moreover, Table~\ref{tab:water-space-short-medium-new} presents model performance on the \ourbenchmark~benchmark in the \textbf{Ship Motion}, covering both river and ocean scenarios across varying reasoning types and difficulty levels. GPT-5 model consistently outperforms other models across all settings.


\begin{table*}[ht]
\centering
\renewcommand{\arraystretch}{1.1}
\caption{Evaluation of \ourbenchmark~in the \textbf{Airplane Navigation} domain using \textbf{Short}, \textbf{Medium}, and \textbf{Long} Videos, categorized by reasoning types, based on a subset of the dataset.}
\Large{
	\begin{adjustbox}{width=1.0\textwidth,center}
\begin{tabular}{llcc|cccc|cccc|cccc}
\toprule
\multirow{2}{*}{\textbf{Difficulty}}  & \multirow{2}{*}{\textbf{Models}}  & \multirow{2}{*}{ \textbf{Size}} &\multirow{2}{*}{ \textbf{Over. Avg.}}  
& \multicolumn{4}{c|}{\textbf{Short Video Scenarios}} & \multicolumn{4}{c|}{\textbf{Medium Video Scenarios}} & \multicolumn{4}{c}{\textbf{Long Video Scenarios}}  \\
 & & &
& \textbf{Avg.} & \textbf{Temporal} & \textbf{Spatial} & \textbf{Intent} & \textbf{Avg.} & \textbf{Temporal} & \textbf{Spatial} & \textbf{Intent} & \textbf{Avg.} & \textbf{Temporal} & \textbf{Spatial} & \textbf{Intent} \\
\midrule 
& GPT 5 \citep{chatgpt5-2025openai}  & -   & {28.11}  
& {26.67} & 18.00 & 30.00 & 32.00  
& {26.00} & 32.00 & 34.00 & 12.00  
& {31.67} & 30.00 & 20.00 & 45.00 \\

    & GPT 4o \citep{hurst2024gpt} & -   & 18.11  & 21.33 & 16.00 & 26.00 & 22.00 & 14.67 & 12.00 & 30.00 & 2.00 & 18.33 & 5.00 & 35.00 & 15.00 \\ 
    & Gemini 2.5 pro  \citep{gemini2dot5-2025}                  & -   & \textbf{31.39} & {32.83} & 36.00 & 24.49 & 38.00 & {24.67}  & 32.00 & 22.00 & 20.00 & {36.67} & 30.00 & 15.00 & 65.00 \\
    & Gemini 2.5 flash think  \citep{gemini2025flash}                    & -   & 25.78 & 26.00 & 26.00 & 18.00 & 34.00 & 21.33 & 28.00 & 18.00 & 18.00 & 30.00 & 30.00 & 10.00 & 50.00 \\
    & Gemini 2.5 flash no-think  \citep{gemini2025flash}                    & -   & 25.44 & 25.33 & 22.00 & 28.00 & 26.00 & 26.00 & 26.00 & 28.00 & 24.00 & 25.00 & 0.00 & 40.00 & 35.00\\
    & Gemini 1.5 pro \citep{google2024gemini15}   & -   & {22.34}  & 26.67 & 24.00 & 26.00 & 30.00 & 18.67 & 20.00 & 22.00 & 14.00 & 21.67 & 10.00 & 25.00 & 30.00 \\
    & Claude 3.5  \citep{anthropic-claude-2025} & -   & 24.22 & 26.00 & 18.00 & 32.00 & 28.00 & 23.33 & 20.00 & 28.00 & 22.00 & 23.33 & 10.00 & 40.00 & 20.0\\
    Hard   & InternVL2.5  \citep{chen2024expanding}                        & 26B  & 17.33 & 19.33 & 24.00 & 26.00 & 10.00 & 19.33 & 16.00 & 32.00 & 10.00 & 13.33 & 10.00 & 10.00 & 20.00 \\
    & InternVL2.5 \citep{chen2024expanding}                         & 8B   & 18.22 & 18.67 & 20.00 & 28.00 & 8.00 & 19.33 & 16.00 & 30.00 & 12.00 & 16.67 & 5.00 & 35.00 & 10.00 \\
     & InternVL2.5 \citep{chen2024expanding}                         & 4B   & 15.33 & 15.33 & 14.00 & 10.00 & 22.00 & 14.00 & 16.00 & 18.00 & 8.00 & 16.67 & 15.00 & 30.00 & 5.00 \\
    & LLaVA Next~\citep{li2024llavanext-ablations} & 32B & 17.89 & 18.67&14.0&34.0&8.00 & 16.67 &6.00 &32.00 &12.00 & 18.33&5.0&40.0&10.00 \\
    & LLaVA Video~\citep{zhang2024video}  & 7B   & 14.78 & 16.67 & 14.00 & 28.00 & 8.00 & 12.67 & 6.00 & 22.00 & 10.00 & 15.00 & 5.00 & 30.00 & 10.00 \\
    & LLaVA OneVision~\citep{li2024llava} & 7B   & 15.67 & 16.00 & 12.00 & 28.00 & 8.00 & 16.00 & 12.00 & 26.00 & 10.00 & 15.00 & 10.00 & 25.00 & 10.00 \\
    & Qwen2.5 VL~\citep{Qwen2.5_VL}       & 32B  & 16.22 & 20.00 & 6.00 & 36.00 & 18.00 & 15.33 & 4.00 & 24.00 & 18.00 & 13.33 & 0.00 & 30.00 & 10.00 \\
    & Qwen2.5 VL~\citep{Qwen2.5_VL}       & 7B   & 16.55 & 19.33 & 0.00 & 30.00 & 28.00 & 15.33 & 2.00 & 30.00 & 14.00 & 15.00 & 5.00 & 30.00 & 10.00 \\

  \midrule
 & GPT 5 \citep{chatgpt5-2025openai}  & -   & {44.00}  
& {39.33} & 28.00 & 44.00 & 46.00  
& {39.33} & 36.00 & 54.00 & 28.00  
& {53.33} & 65.00 & 35.00 & 60.00 \\

    & GPT 4o \citep{hurst2024gpt}  & -   & 38.45  & 38.67 & 38.00 & 56.00 & 22.00 & 30.00 & 38.00 & 34.00 & 18.00 & 46.67 & 65.00 & 30.00 & 45.00  \\
    & Gemini 2.5 pro  \citep{gemini2dot5-2025}                  & -   & {43.11} & {44.67} & 42.00 & 40.00 & 52.00 & {31.33} & 34.00 & 34.00 & 26.00 & {53.33} & 60.00 & 35.00 & 65.0\\
     & Gemini 2.5 flash think  \citep{gemini2025flash}                    & -   & 39.78 & 39.33 & 32.00 & 38.00 & 48.00 & 30.00 & 34.00 & 28.00 & 28.00 & 50.00 & 65.00 & 15.00 & 70.00\\
     & Gemini 2.5 flash no-think  \citep{gemini2025flash}                    & -   & \textbf{49.67} & 43.33 & 30.00 & 48.00 & 52.00 & 40.67 & 38.00 & 50.00 & 34.00 & 65.00 & 60.00 & 65.00 & 70.00\\
    & Gemini 1.5 pro \citep{google2024gemini15}  & -   & {38.78}  & 38.00 & 32.00 & 48.00 & 34.00 & 36.67 & 34.00 & 52.00 & 24.00 & 41.67 & 30.00   & 55.00  & 40.00 \\
    & Claude 3.5  \citep{anthropic-claude-2025} & -   & 39.67 & 38.00 & 26.00 & 40.00 & 48.00 & 36.00 & 32.00 & 54.00 & 22.00 & 45.00 & 50.00 & 35.00 & 50.0\\
    Medium & InternVL2.5  \citep{chen2024expanding}                        & 26B  & 28.67 & 31.33 & 28.00 & 58.00 & 8.00 & 24.67 & 12.00 & 50.00 & 12.00 & 30.00 & 25.00 & 45.00 & 20.00 \\
    & InternVL2.5 \citep{chen2024expanding}                         & 8B   & 34.33 & 30.00 & 20.00 & 58.00 & 12.00 & 34.67 & 32.00 & 50.00 & 22.00 & 38.33 & 40.00 & 45.00 & 30.00 \\
     & InternVL2.5  \citep{chen2024expanding}                        & 4B   & 32.22 & 29.33 & 28.00 & 44.00 & 16.00 & 34.00 & 30.00 & 54.00 & 18.00 & 33.33 & 35.00 & 40.00 & 25.00 \\
    & LLaVA Next~\citep{li2024llavanext-ablations} & 32B & 26.11 & 24.67&18.0&40.0&16.00 & 25.33&18.0&40.0&18.00 &28.33&25.0&40.0&20.00 \\
    & LLaVA Video~\citep{zhang2024video}  & 7B   & 24.00 & 25.33 & 24.00 & 36.00 & 16.00 & 20.00 & 16.00 & 26.00 & 18.00 & 26.67 & 15.00 & 45.00 & 20.00 \\
    & LLaVA OneVision~\citep{li2024llava} & 7B   & 23.67 & 23.33 & 20.00 & 34.00 & 16.00 & 22.67 & 20.00 & 32.00 & 16.00 & 25.00 & 20.00 & 35.00 & 20.00 \\
    & Qwen2.5 VL~\citep{Qwen2.5_VL}       & 32B  & 33.34 & 32.67 & 12.00 & 48.00 & 38.00 & 30.67 & 22.00 & 50.00 & 20.00 & 36.67 & 20.00 & 60.00 & 30.00 \\
    & Qwen2.5 VL~\citep{Qwen2.5_VL}       & 7B   & 28.00 & 24.67 & 16.00 & 24.00 & 34.00 & 26.00 & 24.00 & 26.00 & 28.00 & 33.33 & 35.00 & 20.00 & 45.00 \\

\midrule
& GPT 5 \citep{chatgpt5-2025openai}  & -   & {52.00}  
& {47.33} & 42.00 & 42.00 & 58.00  
& {48.67} & 46.00 & 46.00 & 54.00  
& {60.00} & 65.00 & 35.00 & 80.00 \\

    & GPT 4o \citep{hurst2024gpt}  & -   & 40.67  & 35.33 & 30.00 & 28.00 & 48.00 & 36.67 & 24.00 & 38.00 & 48.00 & 50.00 & 45.00 & 50.00 & 55.00 \\
    & Gemini 2.5 pro  \citep{gemini2dot5-2025}                  & -   & \textbf{52.56} & {56.00} & 60.00 & 48.00 & 60.00 & {40.00} & 40.00 & 36.00 & 44.00 & {61.67} & 75.00 & 35.00 & 75.0\\
     & Gemini 2.5 flash think  \citep{gemini2025flash}                    & -   & 50.67 & 49.33 & 40.00 & 46.00 & 62.00 & 46.00 & 46.00 & 44.00 & 48.00 & 56.67 & 55.00 & 40.00 & 75.00\\
     & Gemini 2.5 flash no-think  \citep{gemini2025flash}                    & -   & 50.78 & 49.33 & 36.00 & 52.00 & 60.00 & 48.00 & 40.00 & 50.00 & 54.00 & 55.00 & 60.00 & 50.00 & 55.00 \\
    & Gemini 1.5 pro \citep{google2024gemini15} & - & 43.00  & 45.33 & 36.00 & 44.00 & 56.00 & 42.00 & 48.00 & 32.00 & 46.00 & 41.67 & 35.00   & 50.00  & 40.00 \\
    & Claude 3.5  \citep{anthropic-claude-2025} & -   & 42.45 & 38.00 & 34.00 & 38.00 & 42.00 & 42.67 & 30.00 & 56.00 & 42.00 & 46.67 & 40.00 & 45.00 & 55.0\\
    Easy & InternVL2.5   \citep{chen2024expanding}                       & 26B  & 36.11 & 35.33 & 36.00 & 44.00 & 26.00 & 34.67 & 28.00 & 46.00 & 30.00 & 38.33 & 30.00 & 40.00 & 45.00 \\
    & InternVL2.5   \citep{chen2024expanding}                       & 8B   & 38.44 & 36.67 & 28.00 & 46.00 & 36.00 & 35.33 & 32.00 & 42.00 & 32.00 & 43.33 & 60.00 & 40.00 & 30.00 \\
    & InternVL2.5   \citep{chen2024expanding}                       & 4B   & 40.33 & 43.33 & 42.00 & 50.00 & 38.00 & 39.33 & 30.00 & 44.00 & 44.00 & 38.33 & 35.00 & 60.00 & 20.00 \\
    & LLaVA Next~\citep{li2024llavanext-ablations} & 32B & 33.22 & 36.67 & 36.00 & 42.00 & 32.00 & 31.33 & 36.00 & 32.00 & 26.00 & 31.67 & 35.00 & 30.00 & 30.00 \\
    & LLaVA Video~\citep{zhang2024video}  & 7B   & 33.22 & 33.33 & 34.00 & 38.00 & 28.00 & 34.67 & 34.00 & 38.00 & 32.00 & 31.67 & 35.00 & 30.00 & 30.00 \\
    & LLaVA OneVision~\citep{li2024llava} & 7B   & 33.22 & 33.33 & 34.00 & 38.00 & 28.00 & 34.67 & 34.00 & 38.00 & 32.00 & 31.67 & 35.00 & 30.00 & 30.00 \\
    & Qwen2.5 VL~\citep{Qwen2.5_VL}       & 32B  & {52.45} & 50.00 & 34.00 & 56.00 & 60.00 & 50.67 & 40.00 & 54.00 & 58.00 & 56.67 & 55.00 & 60.00 & 55.00 \\
    & Qwen2.5 VL~\citep{Qwen2.5_VL}       & 7B   & 39.89 & 33.33 & 28.00 & 18.00 & 54.00 & 38.00 & 48.00 & 16.00 & 50.00 & 48.33 & 55.00 & 30.00 & 60.00 \\

\bottomrule
\end{tabular}
\end{adjustbox}
}
\label{tab:air-space-short-medium-long}
\vspace{-15pt}
\end{table*}

\begin{table*}[ht]
\centering
\renewcommand{\arraystretch}{1.1}
\caption{Evaluation of \ourbenchmark~in the \textbf{Ship Motion} using \textbf{River} and \textbf{Ocean} Videos, categorized by reasoning types, based on a subset of the dataset.}
\Large{
	\begin{adjustbox}{width=0.9\textwidth,center}
\begin{tabular}{llcc|cccc|cccc}
\toprule
\multirow{2}{*}{\textbf{Difficulty}}  & \multirow{2}{*}{\textbf{Models}}  & \multirow{2}{*}{ \textbf{Size}} &\multirow{2}{*}{ \textbf{Over. Avg.}}  
& \multicolumn{4}{c|}{\textbf{River Scenarios}} & \multicolumn{4}{c}{\textbf{Ocean Scenarios}}  \\
 & & &
& \textbf{Avg.} & \textbf{Temporal} & \textbf{Spatial} & \textbf{Intent} & \textbf{Avg.} & \textbf{Temporal} & \textbf{Spatial} & \textbf{Intent} \\
\midrule 
& GPT 5 \citep{chatgpt5-2025openai}  & -   & \textbf{38.36}  
& {48.72} & 46.15 & 30.77 & 69.23  
& {28.00} & 30.00 & 32.00 & 22.00 \\
& GPT4o \citep{hurst2024gpt}  & -   & 22.10  & 28.20 & 38.46 & 26.92 & 19.23 & 16.00 & 18.00 & 18.00 & 12.00 \\
    & Gemini 2.5 pro  \citep{gemini2dot5-2025}                  & -   & {29.64} & {34.62} & 23.08 & 34.62 & 46.15 & {24.67}  & 38.00 & 16.00 & 20.00 \\
     & Gemini 2.5 flash think  \citep{gemini2025flash}                    & -   & 27.36 & 32.05 & 30.77 & 26.92 & 38.46 & 22.67 & 30.00 & 22.00 & 16.00  \\
     & Gemini 2.5 flash no-think  \citep{gemini2025flash}                    & -   & 27.44 & 28.21 & 42.31 & 19.23 & 23.08 & 26.67 & 36.00 & 20.00 & 24.00  \\
    & Gemini 1.5 pro \citep{google2024gemini15}  & -   & {26.02}  & 26.92 & 23.08 & 30.77 & 26.92 & {25.11} & 34.00 & 20.93 & 20.41 \\
    & Claude 3.5 \citep{anthropic-claude-2025} & -   & 25.44 & 28.20 & 19.23 & 19.23 & 46.15 & 22.67 & 26.00 & 22.00 & 20.00 \\
    Hard   & InternVL2.5  \citep{chen2024expanding}      & 26B  & 22.54 & 23.08 & 15.38 & 19.23 & 34.62 & 22.00 & 18.00 & 28.00 & 20.00 \\
    & InternVL2.5  \citep{chen2024expanding}      & 8B   &  21.90 & 21.79 & 7.69  & 26.92 & 30.77 & 22.00 & 16.00 & 28.00 & 22.00 \\
    & InternVL2.5  \citep{chen2024expanding}      & 4B   & 20.92 & 20.51 & 19.23 & 19.23 & 23.08 & 21.33 & 16.00 & 26.00 & 22.00 \\
    & LLaVA Next~\citep{li2024llavanext-ablations} & 32B & 14.39 & 11.54 & 7.69  & 19.23 & 7.69  &    15.33   &   8.00    &   30.00     &  8.00     \\
    & LLaVA Video~\citep{zhang2024video}  & 7B   & 14.00 & 16.67 & 15.38 & 23.08 & 11.54 & 11.33 & 8.00  & 20.00 & 6.00 \\
    & LLaVA OneVision~\citep{li2024llava} & 7B   & 15.67 & 16.67 & 11.54 & 26.92 & 11.54 & 14.67 & 8.00  & 28.00 & 8.00 \\
    & Qwen2.5 VL~\citep{Qwen2.5_VL}       & 32B  & 13.39 & 14.10 & 7.69  & 23.08 & 11.54 & 12.67& 8.0& 24.0& 6.00    \\
    & Qwen2.5 VL~\citep{Qwen2.5_VL}       & 7B   & 14.67 & 16.67 & 7.69  & 30.77 & 11.54 & 12.67 & 6.00  & 24.00 & 8.00 \\

\midrule
& GPT 5 \citep{chatgpt5-2025openai}  & -   & \textbf{51.80}  
& {60.26} & 53.85 & 46.15 & 80.77  
& {43.33} & 56.00 & 48.00 & 26.00 \\
    & GPT 4o \citep{hurst2024gpt}   & -   & 38.49  & 42.31 & 50.00 & 53.85 & 23.08 & 34.67 & 36.00 & 48.00 & 20.00  \\
    & Gemini 2.5 pro  \citep{gemini2dot5-2025}                  & -   & 41.77 & {44.87} & 30.77 & 61.54 & 42.31 & 38.67 & 48.00 & 46.00 & 22.00 \\
     & Gemini 2.5 flash think  \citep{gemini2025flash}                    & -   & {48.26} & 53.85 & 61.54 & 57.70 & 42.31 & 42.67 & 52.00 & 42.00 & 34.00  \\
     & Gemini 2.5 flash no-think  \citep{gemini2025flash}                    & -   & 46.12 & 50.00 & 46.15 & 57.69 & 46.15 & 42.00 & 56.00 & 44.00 & 26.00  \\
    & Gemini 1.5 pro \citep{google2024gemini15} & -   & {46.31} & {53.84}  & 46.15 & 65.38 & 50.00 & 38.78 & 34.00 & 49.02 & 33.33 \\
    & Claude 3.5   \citep{anthropic-claude-2025} & -   & 38.62 & 35.90 & 34.62 & 50.00 & 23.08 & {41.33} & 42.00 & 54.00 & 28.00 \\
    Medium  & InternVL2.5 \citep{chen2024expanding}       & 26B  & 41.77 & 44.87 & 30.77 & 57.69 & 46.15 & 38.67 & 24.00 & 62.00 & 30.00 \\
    & InternVL2.5  \citep{chen2024expanding}       & 8B   & 41.08 & 46.15 & 34.62 & 61.54 & 42.31 & 36.00 & 34.00 & 60.00 & 14.00 \\
     & InternVL2.5 \citep{chen2024expanding}     & 4B   & 44.36 & 48.72 & 23.08 & 65.38 & 57.69 & 40.00 & 28.00 & 60.00 & 32.00 \\
    & LLaVA Next~\citep{li2024llavanext-ablations} & 32B & 20.88 & 23.08 & 11.54 & 38.46 & 19.23 & 18.67 & 10.00 & 30.00 & 16.00 \\
    & LLaVA Video~\citep{zhang2024video}  & 7B   & 21.92 & 20.51 & 19.23 & 26.92 & 15.38 & 23.33 & 20.00 & 30.00 & 20.00 \\
    & LLaVA OneVision~\citep{li2024llava} & 7B   & 22.54 & 23.08 & 19.23 & 30.77 & 19.23 & 22.00 & 14.00 & 34.00 & 18.00 \\
    & Qwen2.5 VL~\citep{Qwen2.5_VL}       & 32B  & 33.31 & 34.62 & 19.23 & 50.00 & 34.62 & 32.00 & 20.00 & 50.00 & 26.00 \\
    & Qwen2.5 VL~\citep{Qwen2.5_VL}       & 7B   & 24.08 & 29.49 & 19.23 & 30.77 & 38.46 & 18.67 & 18.00 & 26.00 & 12.00 \\

\midrule
& GPT 5 \citep{chatgpt5-2025openai}  & -   & \textbf{63.00}  
& {66.67} & 61.54 & 50.00 & 88.46  
& {59.33} & 78.00 & 48.00 & 52.00 \\
    & GPT 4o \citep{hurst2024gpt}   & -   & 50.51  & 57.69 & 57.69 & 50.00 & 65.38 & 43.33 & 66.00 & 34.00 & 30.00\\
    & Gemini 2.5 pro  \citep{gemini2dot5-2025}                  & -   & {61.05} & {64.10} & 57.69 & 57.69 & 76.92 & {58.00} & 72.00 & 50.00 & 52.00 \\
     & Gemini 2.5 flash think  \citep{gemini2025flash}                    & -   & {62.03} & 65.39 & 80.77 & 42.31 & 73.08 & 58.67 & 70.00 & 52.00 & 54.00  \\
     & Gemini 2.5 flash no-think  \citep{gemini2025flash}                    & -   & 58.18 & 57.69 & 57.69 & 38.46 & 76.92 & 58.67 & 80.00 & 42.00 & 54.00  \\
    & Gemini 1.5 pro \citep{google2024gemini15} & -   & 50.69 & 52.56  & 42.31 & 61.54 & 53.85 & 48.81 & 50.00 & 46.43 & 50.00   \\
    & Claude 3.5   \citep{anthropic-claude-2025} & -   & 49.39 & 47.44 & 50.00 & 53.85 & 38.46 & 51.33 & 62.00 & 52.00 & 40.00 \\
   Easy   & InternVL2.5  \citep{chen2024expanding}     & 26B  & {55.05} & 64.10 & 65.38 & 57.69 & 69.23 & 46.00 & 50.00 & 50.00 & 38.00 \\
    & InternVL2.5   \citep{chen2024expanding}   & 8B   & 53.47 & 60.26 & 69.23 & 46.15 & 65.38 & 46.67 & 46.00 & 54.00 & 40.00 \\
     & InternVL2.5  \citep{chen2024expanding}      & 4B   & 53.87 & 56.41 & 53.85 & 57.69 & 57.69 & 51.33 & 52.00 & 56.00 & 46.00 \\
    & LLaVA Next~\citep{li2024llavanext-ablations} & 32B & 35.59 & 37.18 & 26.92 & 53.85 & 30.77 & 34.00 & 30.00 & 38.00 & 34.00 \\
    & LLaVA Video~\citep{zhang2024video}  & 7B   & 31.03 & 32.05 & 30.77 & 34.62 & 30.77 & 30.00 & 22.00 & 38.00 & 30.00 \\
    & LLaVA OneVision~\citep{li2024llava} & 7B   & 33.00 & 33.33 & 34.62 & 34.62 & 30.77 & 32.67 & 28.00 & 38.00 & 32.00 \\
    & Qwen2.5 VL~\citep{Qwen2.5_VL}       & 32B  & 52.77 & 61.54 & 53.85 & 61.54 & 69.23 & 44.00 & 40.00 & 54.00 & 38.00 \\
    & Qwen2.5 VL~\citep{Qwen2.5_VL}       & 7B   & 31.31 & 34.62 & 38.46 & 19.23 & 46.15 & 28.00 & 36.00 & 22.00 & 26.00 \\

\bottomrule
\end{tabular}
\end{adjustbox}
}
\label{tab:water-space-short-medium-new}
\end{table*}

\section{Annotation and Detailed Examples}
\label{appendix:details-examples}

During data annotation, we first define the question types, then watch each video to design corresponding questions and annotate the answers. We first design the hard-level tasks and label each question with the ground-truth answer. Based on these, we then construct the medium and easy tasks. The primary differences between difficulty levels lie in the number and types of answer choices.  Our dataset contains approximately 2,101 videos and 19,136 question–answer pairs, evenly distributed across three difficulty levels: easy ($\approx$ 6,300 Q\&A pairs), medium ($\approx$ 6,300 Q\&A pairs), and hard ($\approx$ 6,300 Q\&A pairs). The difficulty is determined by both the number and type of answer choices. Hard questions typically include 12 choices for temporal and intent reasoning, and 4 for spatial reasoning, requiring precise selection. Medium questions generally offer 6 choices for temporal and intent reasoning, and 3 for spatial reasoning, often involving interval-based options. Easy questions usually present 3 choices, or 2 for spatial reasoning, and also rely on interval-based distinctions.

\begin{figure}[ht]
    \centering
    \includegraphics[width=1.0\linewidth]{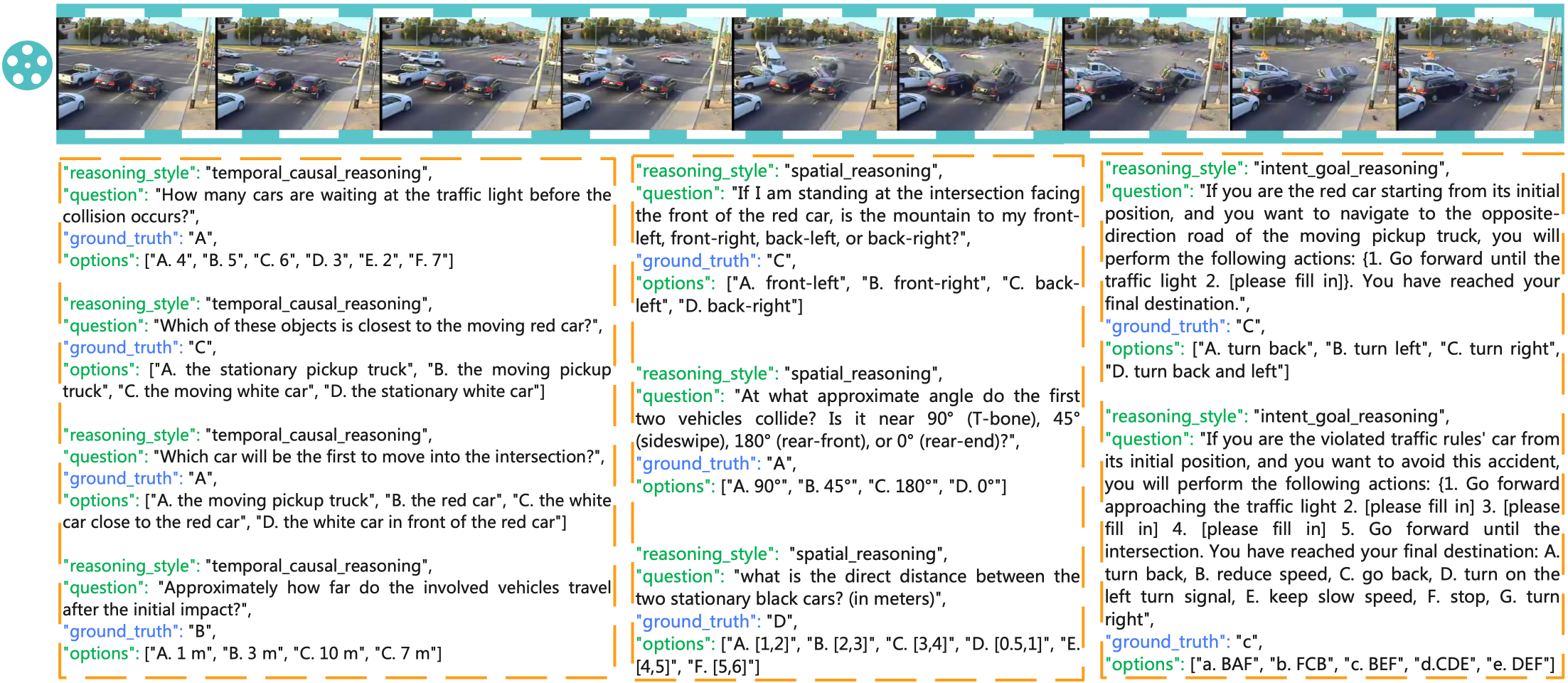}
    \caption{A question and answer example: For each each scenario reasoning setting, we include three types of video lengths: short, medium, and long. Each video length includes tasks designed to evaluate temporal reasoning, spatial reasoning, and intent reasoning.}
    \label{fig:question-answer-template}
\end{figure}

Moreover, as illustrated in Figure~\ref{fig:question-answer-template}, we present a detailed question-and-answer example. For each scenario's understanding and reasoning setting, we include three video lengths, short, medium, and long, each featuring tasks designed to evaluate temporal, spatial, and intent reasoning.

\begin{figure}[tb!]
    \centering
    \begin{minipage}[t]{0.35\linewidth}
        \centering
        \includegraphics[width=\linewidth]{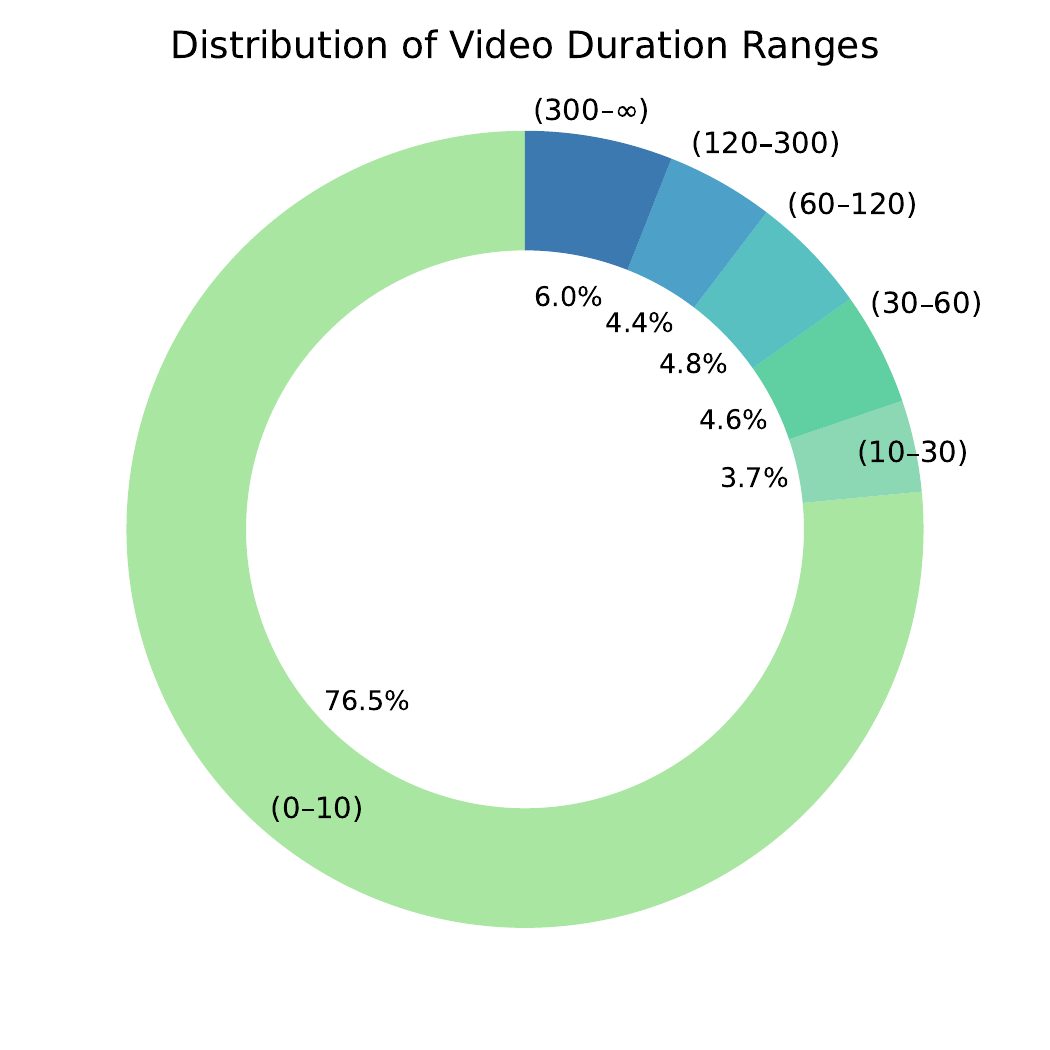}
        \vspace{-20pt}
        
        {(a) Video duration.}
    \end{minipage}
      \hspace{-0.03\linewidth}
    \begin{minipage}[t]{0.35\linewidth}
        \centering
        \includegraphics[width=\linewidth]{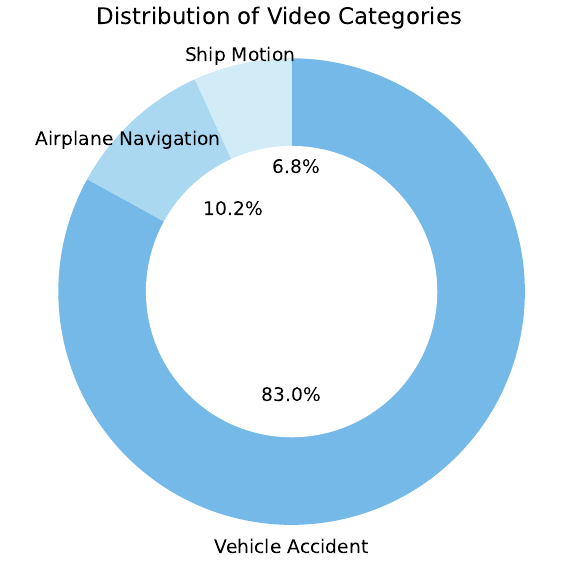} 
        \vspace{-20pt}
        
        {(b) Video categories.}
    \end{minipage}
    \hspace{-0.05\linewidth}
    \begin{minipage}[t]{0.35\linewidth}
        \centering
        \includegraphics[width=\linewidth]{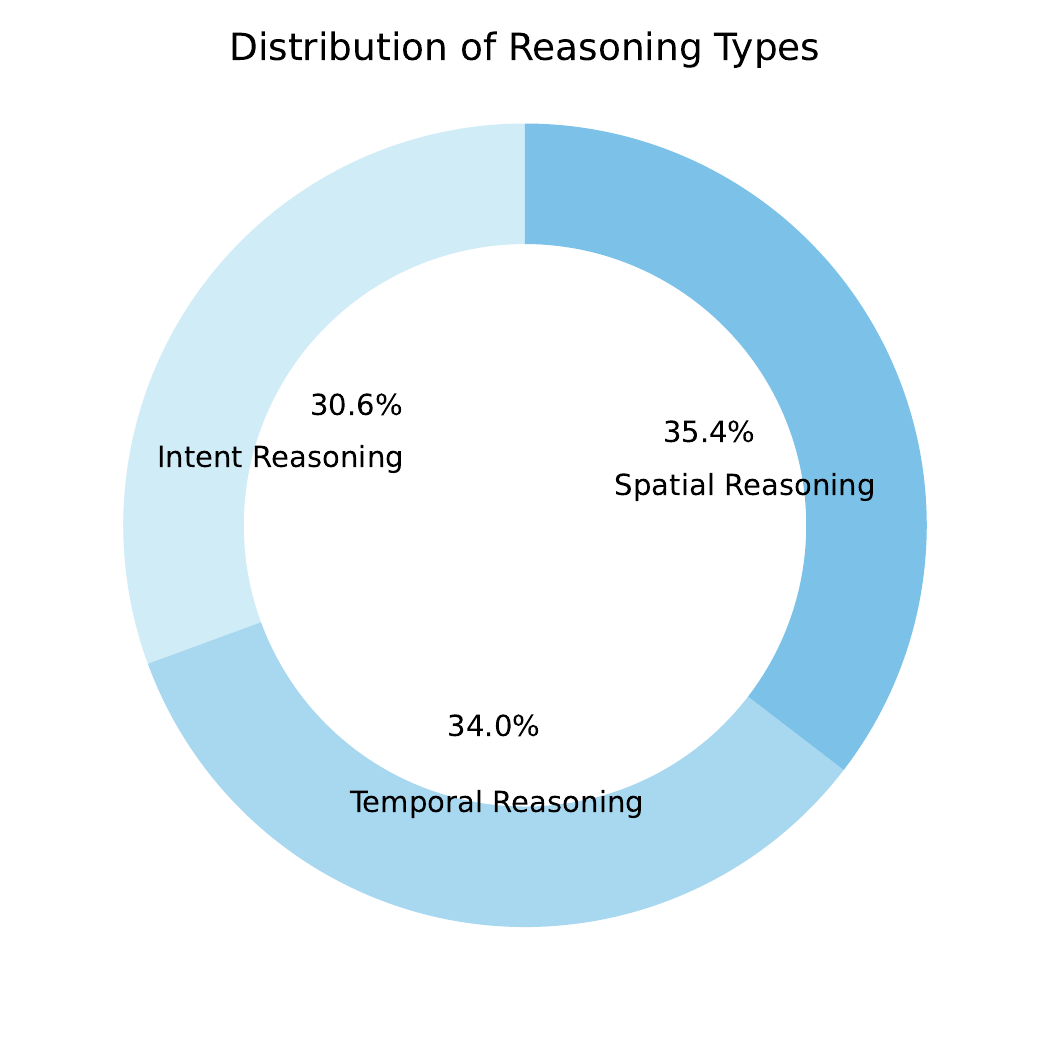}
        \vspace{-20pt}
        
        {(c) Task types.}
    \end{minipage}
    \caption{Distribution of video and task properties in the \ourbenchmark~benchmark.}    
    \label{fig:video-length-frame-reasoning}
    \vspace{-15pt}
\end{figure}

Specifically, as shown in Figure \ref{fig:video-length-frame-reasoning}, in \textbf{(a) Video Length:} A substantial portion of the videos (76.5\%) are short, with durations under 10 seconds. The remaining videos are distributed across longer intervals: 10--30 seconds (3.7\%), 30--60 seconds (4.6\%), 60--120 seconds (4.8\%), 120--300 seconds (4.4\%), and over 300 seconds (6.0\%). This distribution reflects a strong emphasis on short, dynamic scenarios that test rapid perception and reasoning. \textbf{(b) Video Categories:} The benchmark spans three safety-critical domains. {Vehicle Accident}, which primarily involves traffic and safety-related scenarios, comprises 83.0\% of the videos. {airplane navigation} accounts for 10.2\%, and {ship motion} makes up 6.8\%. This distribution highlights both the practical importance of land-based reasoning and the inclusion of underrepresented domains such as maritime and aviation environments. \textbf{(c) Understanding and Reasoning Styles:} \ourbenchmark~supports three major understanding and reasoning types, with a relatively balanced distribution: \textit{spatial reasoning} (35.4\%), \textit{temporal reasoning} (34.0\%), and \textit{intent reasoning} (30.6\%). This design ensures comprehensive evaluation across key dimensions essential for real-world multimodal understanding. Overall, the dataset provides a rich and diverse collection of real-world video scenarios across multiple modalities and time scales, offering a rigorous benchmark for evaluating multimodal understanding and reasoning in safety-critical environments.

\section{Detailed Experiment Settings}

The datasets are used solely for academic research. They are employed only to evaluate model performance and are not used for model training. In our experiments, we build upon the \texttt{lmms-eval} framework~\citep{zhang2024lmmsevalrealitycheckevaluation} as the foundation for our benchmark and extend it to support the specific requirements of \ourbenchmark. All experiments with open-source models were conducted on a Linux system equipped with 8 × NVIDIA A100 GPUs, and experiments with closed-source models were run on a single NVIDIA A100 GPU. Key hyperparameters used for model evaluation are summarized in Table~\ref{appendix-table:evaluation-hyparameters}. More detailed experimental settings are available in our code: \url{https://github.com/SafeRL-Lab/AccidentBench}.

\begin{table}[!htbp]
 \renewcommand{\arraystretch}{1.2}
  \centering
  \caption{Key parameters used in the experiments. }
\label{appendix-table:evaluation-hyparameters}
  \begin{threeparttable}
    \begin{tabular}{cc|cc}
    \toprule
    Parameters & value & Parameters & value \\
    \midrule
    sample size & 1             &       number of processes & 8    \\  
max pixels (Qwen 2.5) & 12845056         & use-flash-attention-2 (Qwen 2.5) & False  \\ 
           interleave visuals (Qwen 2.5)   & True         & enable-chunked-prefill (InternVL 2.5)  & True  \\
           gpu-memory-utilization (InternVL 2.5) &   0.6  & max-num-seqs (InternVL 2.5) & 1  \\
           conv-template (LLava-Video) &  qwen-1-5   & video-decode-backend (LLava-Video) & record  \\
           max-frames-num (LLava-Video) &    22   & mm-spatial-pool-mode (LLava-Video)  & average  \\
           mm-newline-position (LLava-Video) &  grid     & mm-resampler-location (LLava-Video) & after  \\
           conv-template (llava-onevision) & qwen-1-5    & device-map (llava-onevision) & auto  \\
           model-name (llava-onevision) & llava-qwen & &  \\           
    \bottomrule
    \end{tabular}    
    \end{threeparttable}
\end{table}

\end{document}